\documentclass[sigconf]{acmart}
\usepackage{multirow}
\usepackage[table]{xcolor}
\usepackage{array}
\usepackage{tabularx}

\usepackage{booktabs}
\usepackage[table]{xcolor}

\definecolor{bestred}{RGB}{252,225,225}
\definecolor{secondblue}{RGB}{222,237,247}

\renewcommand\footnotetextcopyrightpermission[1]{}

\usepackage[most]{tcolorbox}
\usepackage{xcolor}
\usepackage{needspace}

\definecolor{casegreen}{RGB}{0,128,0}
\definecolor{casered}{RGB}{160,0,0}
\definecolor{casegreenbg}{RGB}{245,252,245}
\definecolor{caseredbg}{RGB}{253,246,246}

\newcommand{\best}[1]{\cellcolor{bestred}#1}
\newcommand{\second}[1]{\cellcolor{secondblue}#1}
\AtBeginDocument{%
  }

\usepackage{xcolor}

\setcopyright{acmlicensed}
\copyrightyear{2018}
\acmYear{2018}
\acmDOI{XXXXXXX.XXXXXXX}
\acmConference[Conference acronym 'XX]{Make sure to enter the correct
  conference title from your rights confirmation email}{June 03--05,
  2018}{Woodstock, NY}
\acmISBN{978-1-4503-XXXX-X/2018/06}

\begin{document}

\title{AUSO: Action-Level Unified Skill Optimization from Internalization to Utilization}

\author{Huizu Lin}
\authornote{Both authors contributed equally to this research.}
\affiliation{%
  \institution{University of Science and Technology of China}
  \city{Hefei}
  \country{China}
}
\email{jordansancholhz@mail.ustc.edu.cn}

\author{Chengkai Huang}
\authornotemark[1]
\affiliation{%
 \institution{University of New South Wales}
 \city{Sydney}
 \country{Australia}}
\email{chengkay.huang@gmail.com}

\author{Tianqi Gao}
\affiliation{%
  \institution{Independent Researcher}
  \country{China}}
\email{tianqig358@gmail.com}
  

\author{Tao Huang}
\authornote{Corresponding author.}
\affiliation{%
  \institution{University of Chinese Academy of Sciences}
  \city{Beijing}
  \country{China}}
\email{thuang@iaii.ac.cn}

\author{Daijiao Liu}
\affiliation{%
  \institution{University of New South Wales}
   \city{Sydney}
 \country{Australia}}
\email{daijiao.liu@student.unsw.edu.au}

\author{Tongxin Li}
\affiliation{%
  \institution{Xi’an Jiaotong-Liverpool University}
  \city{Suzhou}
  \country{China}}
\email{tommy020929@gmail.com}

\author{Xiaoyan Sun}
\affiliation{%
  \institution{University of Science and Technology of China, Hefei Comprehensive National
Science Center
}
  \city{Hefei}
  \country{China}}
\email{sunxiaoyan@ustc.edu.cn}

\author{Lina Yao}
\affiliation{%
  \institution{University of New South Wales}
  \city{Sydney}
  \country{Australia}}
\email{lina.yao@unsw.edu.au}

\renewcommand{\shortauthors}{H. Lin et al.}

\begin{abstract}
Skills play different roles as an agent’s policy evolves: they should first provide learnable knowledge, then support capability formation, and finally be invoked only when they improve individual decisions. Existing methods rarely model this lifecycle. They either keep skills outside the model, fully internalize them, or select among internalization and utilization objectives through noisy task-level success rates. Such designs fragment training and assign uniform importance to actions within the same trajectory, even though skill guidance may help some decisions while distracting others. To solve these problems, we introduce \textbf{AUSO (Action-level Unified Skill Optimization)}, which unifies skill learning and skill use through a progressive, action-aware optimization process. At the beginning of training, AUSO jointly learns from teacher guidance and environmental outcomes, enabling the policy to acquire foundational skills without losing task-oriented feedback. It subsequently emphasizes outcome-based policy optimization to consolidate autonomous problem-solving ability. As the policy matures, AUSO evaluates each sampled action under both skill-conditioned and skill-free contexts. The resulting action-level information signal is coupled with the trajectory outcome advantage, allowing beneficial skill-sensitive actions to receive stronger updates and harmful ones to be suppressed. Therefore, skills gradually transition from an external source of supervision into decision knowledge whose utilization is adapted to its action-level benefit, while reinforcement learning remains the shared backbone across all stages. Experiments on ALFWorld, WebShop, and SearchQA show that AUSO consistently improves agent performance and out-of-distribution generalization over competitive baselines. Our code is available at \underline{\url{https://github.com/JordanSancholhz/Action-Skill}}.
\end{abstract}



\received{20 February 2007}
\received[revised]{12 March 2009}
\received[accepted]{5 June 2009}

\maketitle

\section{Introduction}

Large language models (LLMs) are increasingly evolving from static text generators into interactive agents that make sequential decisions in complex, long-horizon environments. Rather than producing a single response, these agents interleave reasoning, acting, and observation: they call external tools, retrieve information, navigate web interfaces, manipulate embodied environments, and revise plans based on feedback \cite{yao2023react,schick2023toolformer,zhou2023webarena,huang2025towards}. Such interaction-centric settings are central to modern information access and decision-making tasks, including web shopping, embodied household execution, and search-based question answering \cite{yao2022webshop,shridhar2021alfworld,dunn2017searchqa,lou2025speechagent}. However, as task horizons grow, agents must repeatedly decide not only what action to take, but also which procedural knowledge, tool-use strategy, or environmental heuristic should guide that action.

Agent skills have emerged as a promising abstraction for this problem.  They encode reusable procedural knowledge, task strategies, tool-use patterns, or environment-specific guidance, thereby providing agents with transferable priors for long-horizon decision-making \cite{wang2023voyager,huang2026listwise}. In prompt-based agents, skills are retrieved, composed, or injected as explicit guidance to support complex interactive task solving \cite{xia2026skillrl}. 
Recent post-training methods further optimize diverse skill-conditioned behaviors through fine-grained environment feedback, shifting skill usage from static prompts toward learnable components of agent policies \cite{shi2026skill1,zhu2026skill05}.

\begin{figure}
    \centering
    \includegraphics[width=\linewidth]{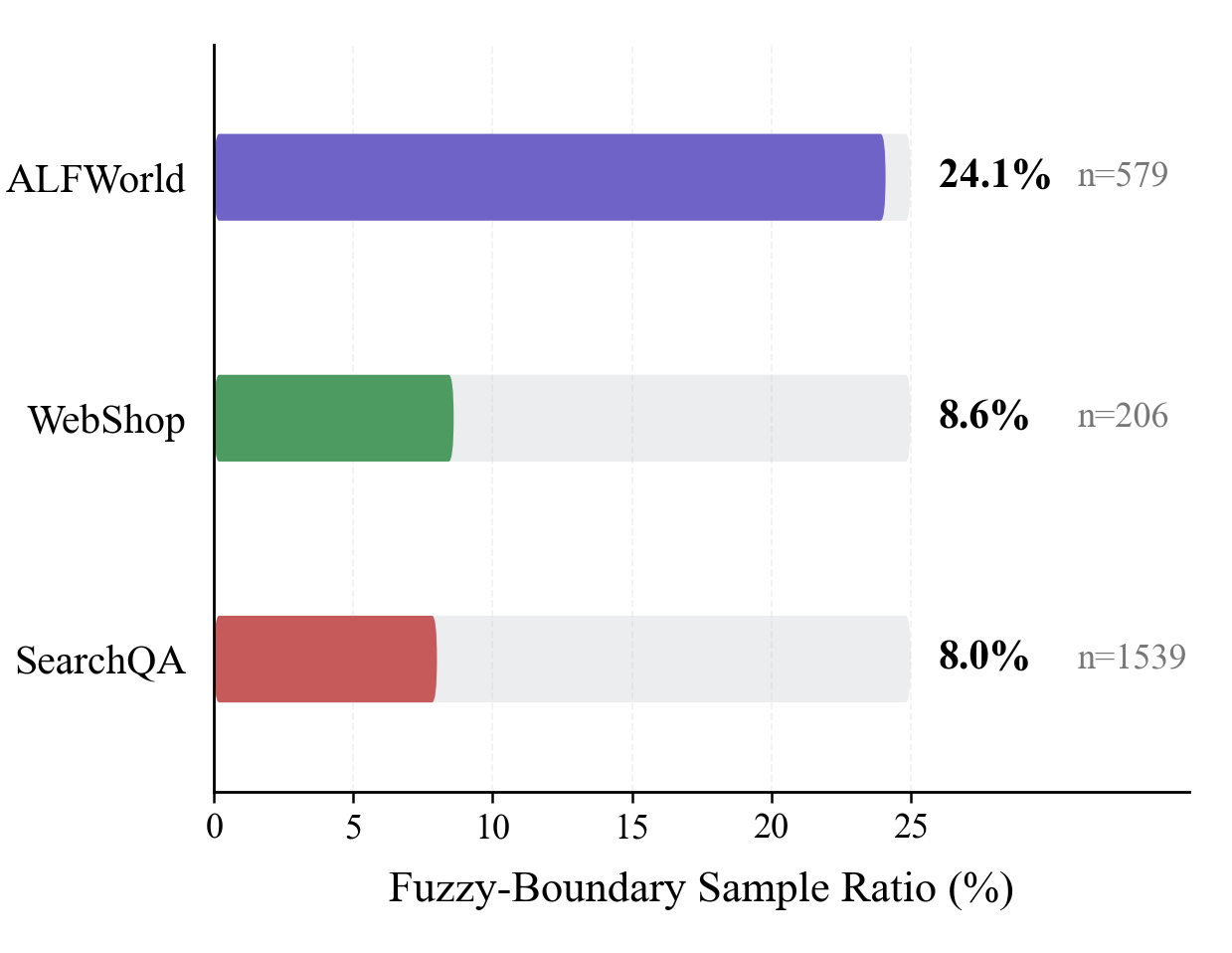}
    \caption{Empirical evidence of ambiguous medium/easy skill-routing boundaries in Skill0.5. Samples across the success threshold can differ by as little as $1/8$, revealing the ambiguity of threshold-based routing.}
    \label{fig:fuzzy-boundary}
\end{figure}

Existing skill-based agent learning methods can be broadly organized into three paradigms. The first paradigm fully externalizes skills: the agent keeps a persistent skill library outside the model and retrieves or updates skills during task execution, as in Voyager \cite{wang2023voyager}, SkillRL \cite{xia2026skillrl}, and Skill1 \cite{shi2026skill1}. This design preserves interpretability, modularity, and compositional reuse, but also introduces context overhead and retrieval noise. The second paradigm fully internalizes skills into the model parameters, aiming to eliminate inference-time skill prompts and enable autonomous execution, as in SKILL0 \cite{lu2026skill0}, LatentSkill \cite{yu2026latentskill}, and related skill-to-parameter approaches \cite{zhang2026skilltolora}. This improves token efficiency and reduces dependence on external memory, but may lose the controllability and task-specific adaptability of explicit skills. The third paradigm jointly externalizes and internalizes skills. Skill0.5 \cite{zhu2026skill05} internalizes generalizable skills into the model while retaining explicit task-specific skill utilization, demonstrating that such a hybrid strategy can better balance cross-task transfer, inference efficiency, and task-level specialization across diverse environments.

Despite this progress, existing hybrid methods still leave a key optimization gap. Skill0.5 relies on task-level success rates to route samples into difficulty tiers and then applies different objectives to hard, medium, and easy tasks. Such routing makes the boundary between internalization and utilization sensitive to noisy aggregate outcomes: samples with very similar empirical success rates may receive different training treatment simply because they fall on opposite sides of a threshold. As shown in Figure~\ref{fig:fuzzy-boundary}, this issue is not rare in practice: across ALFWorld, WebShop, and SearchQA, a substantial number of samples lie around the medium/easy boundary, where their pass rates differ by at most one rollout unit but they are assigned to different optimization objectives. This observation suggests that the inherent boundaries introduced by trajectory-level routing motivate us to move beyond explicit skill selection and pursue a more unified learning paradigm, where skills are first internalized into the policy and then naturally utilized during subsequent interaction. More importantly, trajectory-level skill routing is broad and stable, but may be too coarse for pivotal states where skill guidance changes the quality of a specific decision. In contrast, action-level skill signals are more precise, but are harder to obtain because the environment usually provides only trajectory-level rewards. Such coarse outcome feedback cannot directly reveal which individual actions benefit from skill conditioning and which actions are distracted by it, echoing the credit-assignment challenge in reinforcement learning \cite{schulman2017ppo,shao2024deepseekmath}. As a result, existing methods struggle to optimize skill use at actual granularity of agent decision making both in skills internalization and utilization.

To address these limitations, we propose \textbf{AUSO (Action-level Unified Skill Optimization)}, a progressive reinforcement learning framework that unifies skill internalization and skill utilization through action-aware optimization. At its core, AUSO employs Jensen--Shannon divergence (JSD) as a unified information-gain signal to quantify how skill guidance changes the policy's action distribution, supporting both skill internalization and utilization at the action level. AUSO first allows the policy to learn from both teacher-provided skill guidance and environment outcomes, then consolidates autonomous problem-solving ability through outcome-based policy optimization, and finally evaluates each sampled action under both skill-conditioned and skill-free contexts. By coupling this action-level skill-sensitivity signal with trajectory-level advantage, AUSO strengthens updates for actions that genuinely benefit from skills while suppressing those where skill guidance is harmful. In this way, skills gradually transition from external supervision into decision knowledge utilized according to their action-level benefits, while reinforcement learning remains the shared backbone throughout the agent's policy evolution. Extensive experiments demonstrate that AUSO consistently outperforms strong baselines such as Skill0.5 across WebShop, ALFWorld, and Search-QA, showing robust improvements under both ID and OOD settings on WebShop and ALFWorld, as well as across single-hop and multi-hop tasks on Search-QA which shows effectiveness and generalization ability of AUSO across diverse agent tasks and skill settings. Our contributions are as follows:

\begin{itemize}
    \item We propose AUSO, an Action-level Unified Skill Optimization, which unifies skill internalization and utilization in a progressive reinforcement learning process, enabling skills to evolve from external supervision into decision knowledge utilized according to their action-level benefits as the ability of policy model improves gradually.

    \item We introduce \textbf{action-level skill-sensitive optimization}, which unifies skill internalization and utilization at the action level under a shared JSD-based information-gain measure. It quantifies action-distribution discrepancies induced by skill guidance to facilitate the internalization of generalizable skills and subsequently guide adaptive skill utilization during interaction, overcoming the problem of coarse granularity of trajectory-level skill routing.

    \item We empirically validate our \textbf{AUSO} framework on ALFWorld, WebShop, and SearchQA, showing consistent improvements over competitive baselines in long-horizon agent performance and out-of-distribution generalization.
\end{itemize}

\section{Related Work}

\subsection{LLM Agents}

Large language models have increasingly been studied as interactive agents that solve long-horizon tasks through iterative reasoning, action execution, and feedback from external environments. Prompting methods such as chain-of-thought, ReAct, and Tree-of-Thoughts show that intermediate reasoning and deliberate search can substantially improve complex decision making \cite{wei2022chain,yao2023react,yao2023tree}. Tool-augmented agents further extend LLMs beyond text-only generation by enabling API calls, search, calculation, and other external operations \cite{schick2023toolformer,li2023api}. These capabilities have motivated a broad set of agent benchmarks covering web navigation, online shopping, embodied household tasks, software engineering, and general agent evaluation \cite{yao2022webshop,shridhar2021alfworld,zhou2023webarena,deng2023mind2web,liu2023agentbench,chen2026agent2rlbench}.

Another line of work improves LLM agents through memory \cite{cong2026seeing,ye2026memweaver}, reflection, experience reuse, or reinforcement learning from interaction. Reflexion uses verbal feedback from previous failures to guide future decisions without updating model parameters \cite{shinn2023reflexion}, while Voyager constructs reusable programs and curriculum-driven exploration in open-ended embodied environments \cite{wang2023voyager}. Recent work on agentic reinforcement learning further emphasizes that post-training LLMs for interactive environments differs from standard static preference optimization \cite{huang2025pluralistic,gao2026factorized}, because agents must collect trajectories, interact with environment states, and optimize behavior under delayed outcomes \cite{zhang2025agenticrl,chen2026agent2rlbench}. However, most LLM-agent methods still treat external guidance, memory, and reusable knowledge as auxiliary modules rather than modeling how such knowledge should gradually transition into the policy itself. AUSO addresses this gap by optimizing skill learning and skill use in a unified reinforcement learning cycle process.

\subsection{Agent Skills}

Agent skills provide reusable procedural knowledge for solving complex tasks. In embodied and robotic settings, skills often correspond to executable primitives, affordance-aware subroutines, or programmatic task plans that ground high-level language instructions into feasible actions \cite{ahn2022saycan,singh2023progprompt}. In LLM agents, skills are commonly represented as natural-language strategies, environment-specific heuristics, tool-use patterns, or executable procedures stored in external libraries \cite{wang2023voyager,xia2026skillrl}. This external skill paradigm improves interpretability and modular reuse, but it also introduces retrieval noise, context overhead, and difficulty in deciding which skills should be loaded for a particular decision \cite{jiao2026prunerag,jiao2026doctor,huang2025embedding}.

Recent studies make this bottleneck explicit. Skill Retrieval Augmentation formulates large-scale skill retrieval as a distinct problem and shows that agents often struggle to determine when retrieved skills should actually be incorporated \cite{su2026sra}. SkillComposer studies structured skill composition, where an agent must select not only which skills to use but also how many and in what order \cite{zhao2026skillcomposer}. Skill-usage benchmarking further shows that skill benefits can degrade under realistic retrieval and refinement settings, especially when skills are noisy, mismatched, or not directly tailored to the task \cite{liu2026skillwild}. These works indicate that the external skills are useful but fragile: their effectiveness for dealing with complex tasks depends on accurate retrieval, composition, and selective use.

A complementary direction attempts to internalize skills into model parameters. SKILL0 trains agents to absorb skill knowledge and reduce dependence on explicit skill prompts at inference time \cite{lu2026skill0}, while LatentSkill and Skill-to-LoRA study parameter or latent-space forms of skill internalization \cite{yu2026latentskill,zhang2026skilltolora}. Other recent work studies continual experience internalization and test-time skill evolution, showing that the granularity and timing of experience injection are important for stable long-horizon agent learning \cite{chen2026experience,mao2026lifeskill}. Between full externalization and full internalization, Skill0.5 jointly considers skill utilization and internalization through difficulty-aware routing \cite{zhu2026skill05}. AUSO follows this hybrid motivation, but differs by unifying skill internalization and utilization within a single action-level optimization framework, replacing trajectory-level routing with skill-sensitive learning that progressively internalizes generalizable skills and adapts skill utilization according to its benefit for each action.

\section{Method}

\subsection{Problem Formulation}

We consider a skill-augmented interactive agent that solves long-horizon tasks through multi-step interaction with an environment. 
Each task is formulated as an episodic decision process. 
At time step $t$, the agent observes an interaction history state $h_t=(o_1,a_1,\ldots,o_t)$ and generates an action $a_t \sim \pi_\theta(\cdot \mid h_t)$, where $\pi_\theta$ denotes the policy parameterized by an LLM. 
The environment then returns the next observation $o_{t+1}$, and the episode terminates after $T$ steps or when the task is completed. 
A trajectory is denoted as $\tau=(o_1,a_1,\ldots,o_T,a_T)$, and receives a trajectory-level outcome reward $R(\tau)$, such as task success or task score.

In skill-based agent learning, the policy may additionally condition on a skill $s$, which provides reusable procedural knowledge, task-solving guidance, or environment-specific strategies. 
We denote the skill-conditioned policy calculated as:
\begin{equation}
    \pi_\theta^{+}(a_t \mid h_t, s),
\end{equation}
and the policy without using a skill calculated as:
\begin{equation}
    \pi_\theta^{-}(a_t \mid h_t).
\end{equation}

Existing methods often decide whether a trajectory should be used for skill internalization or skill utilization according to empirical success rates estimated within a sliding window. However, such trajectory-level routing introduces fuzzy decision boundaries and assigns a uniform training role to all actions in the same trajectory, even though skill guidance may be beneficial for some actions but harmful for others. Our goal is therefore to train an agent whose skill learning follows a progressive lifecycle: the policy first internalizes useful skills, then improves through autonomous exploration, and ultimately learns to assess the benefit of skill guidance for each action rather than relying on coarse task-level routing. To realize this objective, we formulate skill-based agent training as a unified progressive policy optimization problem. 
Instead of selecting skill internalization or skill utilization through hard trajectory-level success-rate thresholds, AUSO organizes the learning process into a continuous policy evolution: the agent first internalizes useful skill knowledge, then improves through autonomous exploration, and finally learns to assess the benefit of skill guidance for each action and adapt its utilization accordingly. To support this, AUSO further introduces action-level skill-aware weighting via JSD, so different actions receive different update strengths according to their outcome contribution and dependence on skill guidance.

\subsection{ID and OOD settings}
To evaluate our method under different generalization scenarios, we follow the ID/OOD setting of Skill0.5. Specifically, the task domain space is partitioned into ID domains $\mathcal{D}_{\mathrm{id}}$ and OOD domains $\mathcal{D}_{\mathrm{ood}}$, with their corresponding task sets denoted as $\mathcal{X}^{\mathrm{id}}$ and $\mathcal{X}^{\mathrm{ood}}$. Accordingly, we distinguish the skill space into general skills $\mathcal{S}_{G}$, which capture transferable knowledge shared across domains, and domain-specific skills $\mathcal{S}_{S}$. The latter are further partitioned into ID-specific skills $\mathcal{S}_{S}^{\mathrm{id}}$ and OOD-specific skills $\mathcal{S}_{S}^{\mathrm{ood}}$ according to their associated domains. During training, the agent only interacts with ID tasks $x \sim \mathcal{X}_{\mathrm{train}}^{\mathrm{id}}$, together with the accessible general skills $\mathcal{S}_{G}$ and ID-specific skills $\mathcal{S}_{S}^{\mathrm{id}}$, while OOD tasks $\mathcal{X}^{\mathrm{ood}}$ and their corresponding skills $\mathcal{S}_{S}^{\mathrm{ood}}$ remain unseen. During evaluation, we evaluate the agent on both ID and previously unseen OOD tasks to assess its in-domain performance and cross-domain skill generalization. We adopt this setting for both \textsc{ALFWorld} and \textsc{WebShop}. Since \textsc{Search-QA} does not support the same ID/OOD partition, we follow the original setting of Skill0 for its evaluation.

\subsection{Overview of Action-level Unified Skill Optimization}

\begin{figure*}[t]
    \centering
    \includegraphics[width=\textwidth]{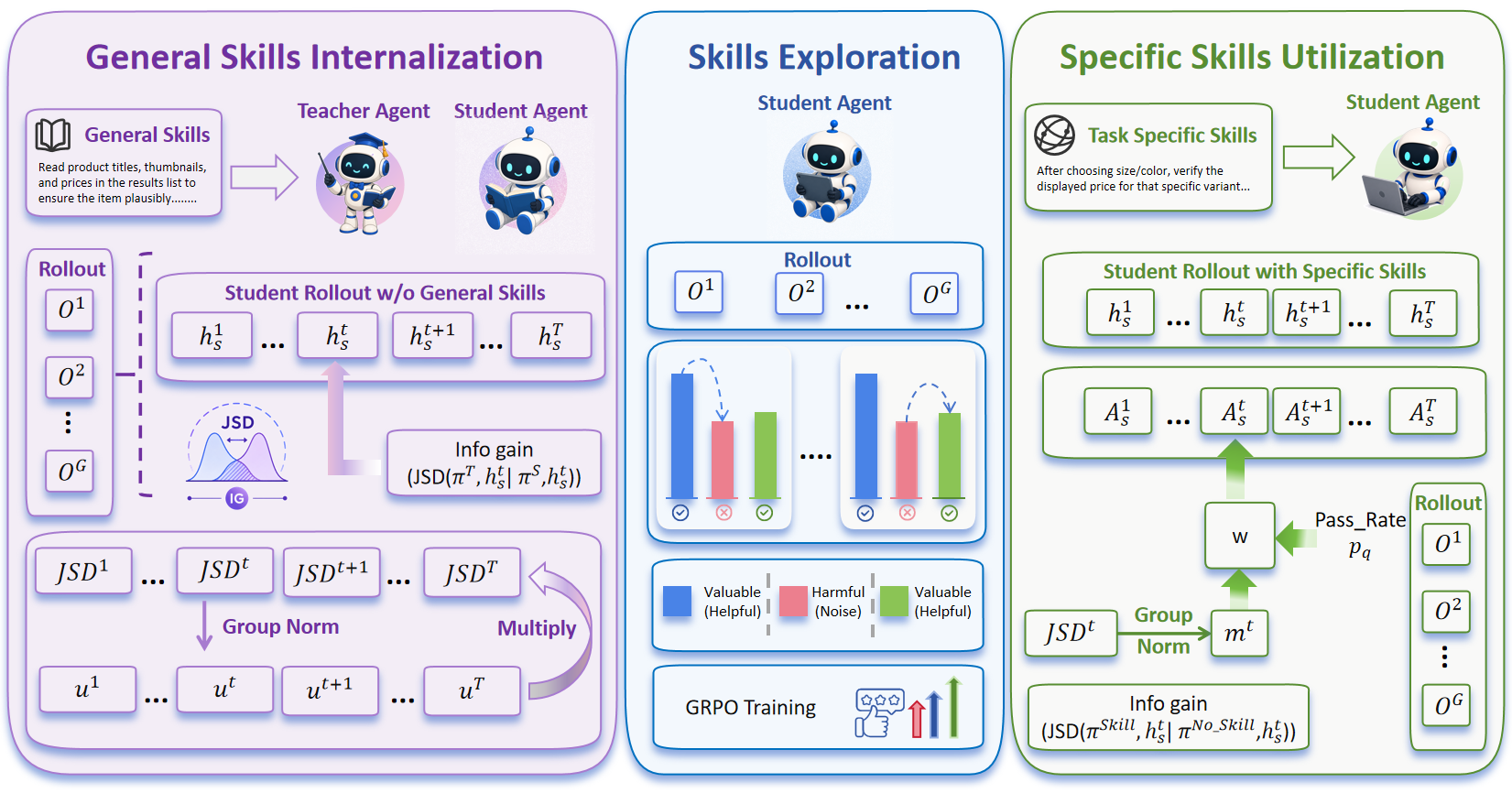}
    \caption{Overview of Our AUSO (Action-Level Unified Skill Optimization from Internalization to Utilization): (1) General Skills Internalization: Internalize transferable general skills through action-level teacher–student guidance. (2) Skills Exploration: Explore and optimize diverse skills via GRPO. (3) Specific Skill Utilization: Evaluate the benefit of task-specific skills for each action to guide action-level skill utilization.
}
    \label{fig:example}
\end{figure*}

To train an agent whose skill learning follows a progressive lifecycle and enables action-level skill-aware optimization, we propose Action-level Unified Skill Optimization (AUSO) framework. Instead of treating skill internalization and skill utilization as two disjoint objectives selected by hard trajectory-level routing, AUSO keeps reinforcement learning via GRPO as the shared optimization backbone and changes how skill information is injected into the update. 
In the early stage, skills provide teacher-guided supervision weighted by action-level information gain \cite{wang2026meg}, so that states where the student deviates more from skill-conditioned teacher behavior receive stronger corrective updates. In the later satge, as the policy becomes more capable through self-exploration, AUSO transitions from externally guided skill internalization to self-guided skill utilization, where the action-level influence of skill conditioning is measured via JSD and used to adaptively reweight policy gradients.


\subsubsection{Early-stage Teacher-guided Skill Internalization}

In the early stage, the main challenge is that outcome-only RL may provide no learning signal for difficult tasks. 
When all sampled trajectories fail, the group-normalized advantage becomes nearly zero according to the advantage of GRPO calculating algorithm:
\begin{equation}
    A_{q,i}^{\mathrm{GRPO}}
    =
    \frac{R_{q,i}-\bar{R}_q}{\sigma_q+\epsilon}
    \approx 0,
    \quad \text{if } R_{q,1}=\cdots=R_{q,G}=0
\end{equation}
where $R_{q,i}$ means the $q^{th}$ trajectory and $i^{th}$ step in the trajectory. In this case, the policy cannot infer which actions should be corrected from the outcome reward alone. 
AUSO therefore introduces a teacher-guided action distribution supervision term, but only for such no-signal task groups.

For each time step t, the agent observes the history state: 
\begin{equation}
    h_t=(o_0,a_0,\ldots,o_t)
\end{equation}

Given the current observation state $h_t$, we aim to internalize skill knowledge into the model rather than relying on explicit skill retrieval during inference. To this end, we construct a teacher--student paradigm from the same model without requiring additional environment rollouts. Specifically, the teacher is conditioned on the available general skills $\mathcal{S}_G$, while the student operates solely on the original observation without access to any skills. Accordingly, we compute two action distributions:

\begin{equation}
\pi_S(\cdot \mid h_t),
\qquad
\pi_T(\cdot \mid h_t, \mathcal{S}_G)
\end{equation}

where $\pi_T$ represents the teacher distribution with $S_{G}$that provides guidance for transferring the generalizable knowledge encoded in $\mathcal{S}_G$, while $\pi_S$ denotes the skill-free student distribution. The teacher distribution is detached during optimization, such that the supervision exclusively updates the student model, progressively internalizing the general skills into its policy parameters.

We measure the discrepancy as the information gain between teacher and student by Jensen-Shannon divergence:
\begin{equation}
    D_{t,k}^{T\rightarrow S}
    =
    \mathrm{JSD}
    \left(
    \pi_T(\cdot \mid h_t, C_{\mathrm{teacher}})
    \,\|\, 
    \pi_S(\cdot \mid h_t)
    \right)
\end{equation}
computed over the top-$k$ teacher tokens. 
Since one environment action may contain multiple generated tokens, we aggregate token-level divergences within the action span $\mathcal{A}_t$:
\begin{equation}
    D_t^{T\rightarrow S}
    =
    \frac{1}{|\mathcal{A}_t|}
    \sum_{k\in \mathcal{A}_t}
    D_{t,k}^{T\rightarrow S}
\end{equation}
This prevents longer textual actions or reasoning tokens from receiving disproportionately large supervision weight.

To make teacher supervision action-aware while keeping the loss scale stable,
AUSO first normalizes the teacher-student discrepancy among valid actions in the same task group:
\begin{equation}
    z_t =
    \frac{D_t^{T\rightarrow S}-\mu_q}
    {\sqrt{\sigma_q^2+\epsilon}}
\end{equation}
where $D_t^{T\rightarrow S}$ denotes the action-level teacher-student divergence at action $t$, and $\mu_q$ and $\sigma_q^2$ are the mean and variance of such divergences over valid actions in task group $q$. We then transform the normalized discrepancy $z_{t}$ into a centered and bounded modulation:
\begin{equation}
    \hat{u}_t =
    \frac{
    \tanh(z_t) - \mathbb{E}_{t\in q}[\tanh(z_t)]
    }
    {
    \max_{t\in q}
    \left|
    \tanh(z_t) - \mathbb{E}_{t\in q}[\tanh(z_t)]
    \right|
    + \epsilon
    } 
\end{equation}
Here, $\tanh(\cdot)$ bounds the modulation, the expectation term recenters the weights within the task group, and the denominator normalizes the maximum magnitude so that $\hat{u}_t \in [-1,1]$.

Finally, the action-level teacher supervision weight is defined as
\begin{equation}
    u_t = 1 + \beta_{\mathrm{int}}\hat{u}_t,
    \qquad
    u_t \in [1-\beta_{\mathrm{int}}, 1+\beta_{\mathrm{int}}]
\end{equation}

Following the stabilization principle of clipped policy optimization \cite{shao2024deepseekmath}, we cap $\beta_{\mathrm{int}}$ with an action-level internalization clip, so that teacher-guided JSD only mildly reweights actions instead of changing the overall optimization scale. 
This prevents high-discrepancy actions from dominating the gradient while still assigning them stronger corrective supervision. The early-stage distillation loss is:
\begin{equation}
    \mathcal{L}_{\mathrm{JSD}}
    =
    \frac{1}{N}
    \sum_t
    u_t D_t^{T\rightarrow S}
\end{equation}

We control the strength of teacher-guided internalization with a time-dependent coefficient:
\begin{equation}
    \lambda_{\mathrm{JSD}}(s)
    =
    \lambda_0 \alpha(s)
\end{equation}
where $\lambda_0$ is the base JSD coefficient and $\alpha(s)$ determines how strongly teacher supervision contributes at training step $s$. 
Following the general principle of curriculum and scheduled supervision, auxiliary guidance should be introduced gradually and annealed as the learner becomes more capable \cite{bengio2009curriculum,bengio2015scheduled}. 
In AUSO, this is implemented with a short ramp-up period followed by a smooth decay: the teacher-guided signal is gradually activated during the first $r$ training steps, and then annealed as the policy shifts toward autonomous exploration. 
We therefore define $\alpha(s)$ as a ramp-up term multiplied by a smooth decay term:
\begin{equation}
    \alpha(s)
    =
    \mathrm{Ramp}(s)
    \cdot
    \left[
    1 -
    \mathrm{SmoothStep}
    \left(
    \frac{s/T}{\rho_{\mathrm{int}}}
    \right)
    \right]
\end{equation}
where $s$ is the current training step, $T$ is the total number of training steps, and $\rho_{\mathrm{int}}$ denotes the fraction of training allocated to the internalization phase. The ramp function is defined as:
\begin{equation}
    \mathrm{Ramp}(s)
    =
    \min\left(1,\frac{s}{r}\right)
\end{equation}
where $r$ is the number of ramp-up steps. 
The smooth decay uses the standard SmoothStep function, a cubic Hermite interpolation commonly used for smooth transitions \cite{perlin1985image,khronos2026smoothstep}:
\begin{equation}
    \mathrm{SmoothStep}(x)
    =
    \tilde{x}^{2}(3-2\tilde{x})
    \qquad
    \tilde{x}=\mathrm{clip}(x,0,1)
\end{equation}

With this schedule, teacher supervision is gradually activated at the beginning of training and then smoothly annealed, so that the policy first receives corrective skill guidance but is later encouraged to rely on environment-driven optimization. 
The early-stage objective is finally written as:
\begin{equation}
    \mathcal{L}_{\mathrm{early}}
    =
    \mathcal{L}_{\mathrm{GRPO}}
    +
    \mathbb{I}[p_q=0]\,
    \lambda_{\mathrm{JSD}}(s)
    \mathcal{L}_{\mathrm{JSD}}
\end{equation}

where $p_{q}$ means the success rate of the student agent in task q.

\begin{table*}[t]
\centering
\caption{Performance comparison on WebShop and ALFWorld under 
in-distribution (ID) and out-of-distribution (OOD) settings.}
\label{tab:webshop_alfworld}
\renewcommand{\arraystretch}{1.08}
\setlength{\tabcolsep}{2.8pt}
\small

\resizebox{\textwidth}{!}{
\begin{tabular}{l|ccccc|cccc|cccc|cccc}
\hline

\multirow{3}{*}{\textbf{Method}}
& \multicolumn{9}{c|}{\textbf{WebShop}}
& \multicolumn{8}{c}{\textbf{ALFWorld}} \\
\cline{2-18}

& \multicolumn{5}{c|}{\textbf{ID}}
& \multicolumn{4}{c|}{\textbf{OOD}}
& \multicolumn{4}{c|}{\textbf{ID}}
& \multicolumn{4}{c}{\textbf{OOD}} \\
\cline{2-18}

& App. & Elec. & Foot. & Other & Avg.
& Access. & Beauty & Home & Avg.
& Pick & Cool & Clean & Avg.
& Look & Heat & Pick2 & Avg. \\

\hline
\multicolumn{18}{l}{\textit{Prompt-based Methods}} \\
\hline

Zero-shot
& 4.4 & 4.6 & 3.4 & 1.0 & 3.5
& 2.5 & 3.7 & 5.5 & 3.9
& 28.6 & 12.0 & 18.5 & 20.7
& 38.5 & 12.5 & 12.5 & 18.9 \\

Few-shot
& 14.2 & 15.1 & 13.5 & 24.0 & 16.5
& 18.8 & 29.6 & 5.5 & 16.9
& 62.9 & 44.0 & 63.0 & 57.5
& 46.2 & 31.2 & 8.3 & 24.5 \\

ReAct
& 12.4 & 11.1 & 4.5 & 12.0 & 10.4
& 11.2 & 24.1 & 2.7 & 11.6
& 71.4 & 28.0 & 33.3 & 47.1
& 46.2 & 18.8 & 12.5 & 22.6 \\

Reflexion
& 3.5 & 8.6 & 1.1 & 7.0 & 5.5
& 6.3 & 5.6 & 1.4 & 4.4
& 85.7 & 44.0 & 44.4 & 60.9
& 46.2 & 31.3 & 29.2 & 34.0 \\

Mem0
& 8.9 & 9.2 & 2.3 & 11.0 & 8.2
& 10.0 & 16.7 & 4.1 & 9.7
& 54.3 & 4.0 & 18.5 & 28.7
& 38.5 & 18.8 & 4.2 & 17.0 \\

ExpeL
& 6.2 & 12.5 & 9.0 & 21.0 & 12.1
& 12.5 & 25.9 & 8.2 & 14.5
& 80.0 & 44.0 & 66.7 & 65.5
& 46.2 & 18.8 & 20.8 & 18.9 \\

MemP
& 15.9 & 13.2 & 9.0 & 19.0 & 14.3
& 16.2 & 14.8 & 9.6 & 13.5
& 65.7 & 12.0 & 33.3 & 40.2
& 46.2 & 37.5 & 12.5 & 28.3 \\

SimpleMem
& 11.5 & 13.8 & 6.7 & 13.0 & 11.7
& 10.0 & 20.4 & 5.5 & 11.1
& 71.4 & 16.0 & 44.4 & 47.1
& 53.8 & 18.8 & 20.8 & 28.3 \\

\hline
\multicolumn{18}{l}{\textit{RL-based Methods}} \\
\hline

RLOO
& 34.9 & 23.9 & \second{41.5} & 32.1 & 31.1
& 31.4 & 46.5 & 22.5 & 32.9
& 91.4 & 80.0 & 81.5 & 85.1
& 61.5 & 56.3 & 20.8 & 41.5 \\

GRPO
& 35.1 & 22.6 & 39.0 & 49.5 & 33.6
& 27.7 & 47.3 & 25.9 & 32.3
& 80.0 & 72.0 & 88.9 & 80.5
& 76.9 & 56.3 & 16.7 & 43.4 \\

\hline
\multicolumn{18}{l}{\textit{Memory-Augmented RL Methods}} \\
\hline

MemRL
& 22.2 & 15.2 & 25.0 & 48.3 & 26.2
& 13.2 & 17.9 & 27.8 & 19.6
& 74.3 & 12.0 & 55.6 & 50.6
& 46.2 & 12.5 & 45.8 & 35.8 \\

EvolveR
& 32.5 & 31.1 & 25.0 & 20.9 & 28.0
& 20.8 & 28.6 & 18.2 & 21.9
& 88.6 & 52.0 & 81.5 & 75.9
& 46.2 & 6.2 & 50.0 & 35.8 \\

Mem0+GRPO
& 36.5 & 22.0 & 40.6 & 23.1 & 29.5
& 10.2 & 32.1 & 29.4 & 25.0
& 65.7 & 20.0 & 51.9 & 48.3
& 23.1 & 6.2 & 20.8 & 17.0 \\

SimpleMem+GRPO
& 25.4 & 28.0 & 26.6 & 25.1 & 26.4
& 16.2 & 32.1 & 29.4 & 25.0
& 85.7 & 52.0 & 70.3 & 71.3
& 61.5 & 43.8 & 41.7 & 47.2 \\

\hline
\multicolumn{18}{l}{\textit{Skill-Augmented RL Methods}} \\
\hline

SkillRL
& 36.0 & 34.2 & 41.4 & 49.3 & 38.1
& 36.3 & 48.5 & 27.6 & 36.7
& 91.4 & 84.0 & \best{96.3} & 90.8
& \best{69.2} & 75.0 & 12.5 & 45.3 \\

Skill0
& \second{39.2} & 33.0 & 38.1 & 37.9 & 35.2
& \second{42.1} & 38.6 & 26.5 & 35.4
& \best{94.3} & 76.0 & 81.5 & 85.1
& 46.2 & 50.0 & 29.2 & 39.6 \\

SLIM
& 31.9 & 36.8 & 31.5 & 33.0 & 33.7
& 35.0 & 29.6 & \second{35.6} & 33.8
& 91.4 & 84.0 & 70.4 & 82.8
& 53.8 & 31.3 & 29.2 & 35.8 \\

Skill0.5
& 39.1 & \second{37.3} & 41.1 & \second{50.9} & \second{40.4}
& 36.6 & \second{54.2} & 31.4 & \second{40.6}
& \best{94.3} & \second{88.0} & \best{96.3} & \second{93.1}
& \best{69.2} & \best{87.5} & \second{33.3} & \second{58.5} \\

\textbf{AUSO (Ours)}
& \best{47.4} & \best{48.0} & \best{50.1} & \best{60.0} & \best{49.7}
& \best{53.1} & \best{62.9} & \best{39.7} & \best{51.2}
& \second{91.4} & \best{96.0} & \best{96.3} & \best{94.3}
& \best{69.2} & \best{87.5} & \best{54.2} & \best{67.9}  \\

\hline
\end{tabular}
}
\end{table*}

\subsubsection{Outcome-driven Autonomous Exploration}

After early skill internalization, AUSO removes teacher-guided supervision and lets the agent improve through standard GRPO. 
In this stage, the policy samples multiple trajectories for each task and uses group-normalized outcome advantages to update itself:
\begin{equation}
    \mathcal{L}_{\mathrm{mid}}=\mathcal{L}_{\mathrm{GRPO}}
\end{equation}
This encourages the agent to consolidate autonomous problem-solving ability through environment-driven exploration.

\subsubsection{Later-stage Action-level Skill Utilization}

In the later stage, AUSO stops using the external teacher as a distillation target. 
The question is no longer how to imitate teacher behavior, but when the student's own decision is meaningfully affected by skill information. 
For every state $H_t$ visited by the student, we compute two distributions from the same policy under two contexts:
\begin{equation}
    \pi_\theta(\cdot \mid h_t, C=\mathrm{Skill}),
    \qquad
    \pi_\theta(\cdot \mid h_t, C=\mathrm{No\text{-}skill})
\end{equation}
Because both distributions are evaluated at the same visited state, this comparison does not require additional environment rollouts and avoids the mismatch that would arise from comparing two separately generated trajectories. Action-level skill information is:
\begin{equation}
    I_t =
    \mathrm{JSD}
    \left(
    \pi_\theta(\cdot \mid h_t, C=\mathrm{Skill})
    \,\|\, 
    \pi_\theta(\cdot \mid h_t, C=\mathrm{No\text{-}skill})
    \right)
\end{equation}
Here, $I_t$ is not used as a distillation loss. 
It is an information measure: a larger value means that skill conditioning changes the student's next-action distribution more strongly at state $H_t$.

Since the raw scale of $I_t$ varies across tasks, AUSO standardizes it within each task group:
\begin{equation}
    z^1_t =
    \frac{I_t-\mu^t_q}{\sqrt{\sigma_{q,I}^2+\epsilon}},
    \qquad
    m_t = \tanh(\mathrm{clip}(z^1_t,-3,3))
\end{equation}
The normalized score $m_t\in[-1,1]$ indicates whether an action is more or less skill-sensitive than the task-level average.

AUSO further modulates the action-level signal by a global uncertainty gate based on the normalized Bernoulli variance:
\begin{equation}
    g(p_q)=K \cdot p_q(1-p_q)
\end{equation}
where K is a positive scaling constant. This form follows the common uncertainty principle that binary outcomes are most informative when their empirical probability is near $0.5$ and least informative when the outcome is almost deterministic \cite{settles2009active}. 
Thus, the gate is small when all rollouts fail or all rollouts succeed, because the group outcome provides little contrast for action-level credit assignment.  We provide a detailed mathematical justification of this gate in Appendix \ref{app:competence_gate}. It reaches its maximum when successful and failed rollouts coexist, where outcome differences are most informative for identifying which skill-sensitive actions matter. The final action weight is calculated as:
\begin{equation}
    w_t = 1 + \beta(s) g(p_q) m_t
\end{equation}

\begin{figure*}[t]
    \centering
    \includegraphics[width=\linewidth]{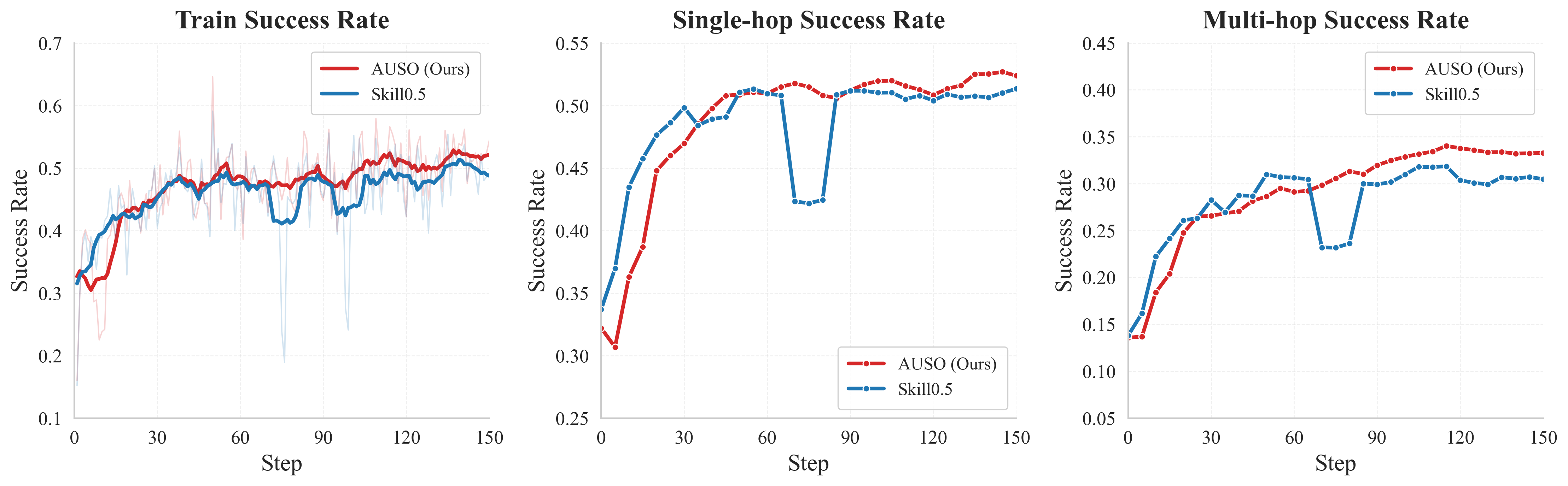}
    \caption{Single-hop and multi-hop validation success rates and training success rates of our method and Skill0.5 with task-specific skills on Search QA benchmarks.}
    \label{fig:search}
\end{figure*}

Same as the curriculum strategy in the early stage, we also introduce $\beta(s)$, which smoothly increases during training:
\begin{equation}
    \beta(s)
    =
    \operatorname{SmoothStep}
    \left(
        \frac{s-s_{\mathrm{start}}}
        {T-s_{\mathrm{start}}}
    \right)
\end{equation}

where $s$ denotes the current training step, $s_{\mathrm{start}}$ is the step at which utilization optimization stage is activated, and $T$ is the total number of training steps. The weighted advantage becomes:
\begin{equation}
    A_{q,t}^{\mathrm{AUSO}}
    =
    A_{q}^{\mathrm{GRPO}} w_t.
\end{equation}
This reweighting preserves the sign of the original RL advantage:
\begin{equation}
    A_q^{\mathrm{GRPO}}>0 \Rightarrow A_{q,t}^{\mathrm{AUSO}}>0,
    \qquad
    A_q^{\mathrm{GRPO}}<0 \Rightarrow A_{q,t}^{\mathrm{AUSO}}<0
\end{equation}
Therefore, AUSO does not create artificial rewards or reverse the trajectory-level outcome signal. 
Instead, it redistributes update strength across actions according to the information gain: in successful trajectories, highly skill-sensitive actions receive stronger positive reinforcement; in failed trajectories, highly skill-sensitive actions receive stronger suppression; and low-information actions receive smaller updates. 
The later-stage objective is still standard on-policy RL, but with action-level skill-aware advantages:
\begin{equation}
    \mathcal{L}_{\mathrm{late}}
    =
    -\sum_t
    A_{q,t}^{\mathrm{AUSO}}
    \log \pi_\theta(a_t\mid h_t, C=\mathrm{Skill})
\end{equation}

\subsubsection{Unified Skill Lifecycle Objective}

The complete AUSO objective is formulated as:
\begin{equation}
\begin{aligned}
\mathcal{L}_{\mathrm{AUSO}}(\theta;s)
={}&
\mathcal{L}_{\mathrm{GRPO}}
\left(
\theta;
A_{q,i}^{\mathrm{GRPO}}
\left[
1+\beta(s)\,4p_q(1-p_q)\,m_{q,i,t}
\right]
\right)
\\
&+
\lambda_0\alpha(s)
\sum_q \mathbb{I}[p_q=0]\,
\mathcal{L}_{\mathrm{JSD}}^{(q)} .
\end{aligned}
\end{equation}
Here, $\alpha(s)$ gradually removes teacher-guided supervision, while $\beta(s)$ progressively activates action-level skill utilization. AUSO retains outcome-driven GRPO as a persistent optimization backbone and progressively advances from external skill internalization to action-level skill utilization guided by skills benefit for each action.
\begin{table}[t]
\centering
\caption{Performance comparison on Search-QA benchmark.}
\label{tab:search_qa}
\renewcommand{\arraystretch}{1.2}
\setlength{\tabcolsep}{2.3pt}
\footnotesize

\resizebox{\columnwidth}{!}{
\begin{tabular}{l|ccc|cccc|c}
\hline
\multirow{2}{*}{\textbf{Method}}
& \multicolumn{3}{c|}{\textbf{Single-hop}}
& \multicolumn{4}{c|}{\textbf{Multi-hop}}
& \multirow{2}{*}{\textbf{Avg.}} \\
\cline{2-8}
& \textbf{NQ}
& \textbf{Triv}
& \textbf{Pop}
& \textbf{Hotp}
& \textbf{2Wk}
& \textbf{MuS}
& \textbf{Bam}
& \\
\hline

Zero-Shot
& 10.4 & 32.4 & 22.3 & 15.8 & 15.4 & 7.2 & 19.2 & 17.5 \\

Few-Shot$^{\dagger}$
& 12.3 & 36.8 & 24.5 & 17.7 & 18.2 & 6.5 & 24.8 & 20.1 \\

Zero-Shot$^{\star}$
& 6.9 & 30.4 & 12.0 & 10.5 & 9.1 & 5.5 & 24.0 & 14.0 \\

Few-Shot$^{\star\dagger}$
& 10.5 & 31.9 & 18.7 & 14.2 & 14.4 & 6.9 & 24.8 & 17.3 \\

GRPO
& 45.1 & \second{63.7} & 44.0 & 43.6
& \best{43.2} & 16.8 & 37.6 & 41.9 \\

OPSD
& 8.8 & 8.6 & 17.5 & 2.5
& 4.2 & 1.2 & 6.2 & 4.5 \\

AgentOCR$^{*}$
& 43.1 & 61.0 & 45.4 & 40.8
& 38.3 & 15.7 & 36.8 & 40.1 \\

EvolveR
& 43.5 & 63.4 & 45.9 & 38.2
& \second{42.0} & 15.6 & 54.4 & 43.1 \\

SkillRL
& \second{45.9} & 63.3 & 45.9 & 43.2
& 40.3 & \best{20.2} & \best{73.8} & \second{47.1} \\

LatentSkill
& 36.2 & 57.6 & 41.0 & 39.6
& 32.0 & 9.80 & 25.6 & 35.6 \\

Skill0
& 42.7 & 61.1 & 45.3 & 40.0
& 38.3 & 16.4 & 66.9 & 44.4 \\

Skill0.5
& 44.7 & 60.4 & 43.8 & 39.0
& 39.2 & 12.0 & 65.3 & 44.2 \\

Skill1
& \best{46.8} & 46.3 & \best{47.8} & \second{43.7}
& 39.3 & \second{18.2} & \second{70.6} & 44.7 \\

\textbf{AUSO (Ours)}
& 45.0 & \best{64.7} & \second{46.1} & \best{44.0}
& 41.8 & 15.4 & 68.1 & \best{47.5} \\

\hline
\end{tabular}
}
\vspace{-7pt}
\end{table}

\section{Experiments}

\subsection{Experiment Settings}

\subsubsection{Datasets}

We evaluate AUSO on three long-horizon agentic benchmarks: ALFWorld \cite{shridhar2021alfworld}, WebShop \cite{yao2022webshop}, and SearchQA \cite{dunn2017searchqa}. 
These benchmarks cover embodied decision-making, web-based interaction, and search-oriented question answering, allowing us to examine whether AUSO can improve both task ID performance and OOD (out-of-distribution) generalization across different forms of agent interacting with different environment.

\textbf{ALFWorld} is an embodied household task benchmark that aligns text-based environments with embodied manipulation scenarios. 
An agent must follow natural-language goals and complete multi-step household tasks through textual actions such as navigation, object interaction, and state manipulation. 
We use ALFWorld to evaluate whether skill guidance can improve long-horizon planning and action execution in embodied environments.

\textbf{WebShop} is a web-based shopping benchmark where an agent interacts with a simulated e-commerce website to find and purchase products satisfying user instructions. 
The agent must search, compare products, inspect attributes, and make purchase decisions through multi-step web actions. 
Following our out-of-distribution setting, we construct category-based ID and OOD splits according to Skill0.5: ID tasks are drawn from frequent categories such as apparel, electronics, footwear, and other products, while OOD validation tasks are drawn from held-out categories such as accessories, beauty and health, and home decor.

\textbf{SearchQA} is a search-based question answering benchmark where an agent must retrieve useful information and produce answers based on external evidence. 
Compared with embodied and web-shopping tasks, SearchQA emphasizes information seeking and evidence aggregation, providing a complementary testbed for evaluating whether AUSO can improve skill use in knowledge-intensive decision processes.

\begin{table}[t]
\centering
\renewcommand{\arraystretch}{1.2}
\caption{Ablation studies of the average success rate on ALFWorld and WebShop to validate the effectiveness of all our core modules; the best performance of each metric is bolded, and the second‑best result is underlined.}
\begin{tabular}{lcc cc}
\hline
Variant & \multicolumn{2}{c}{ALFWorld} & \multicolumn{2}{c}{WebShop} \\
& ID Avg. & OOD Avg. & ID Avg. & OOD Avg. \\
\hline
w/o Int.-Action & 85.1 & \underline{54.7} & 44.7 & \underline{47.3} \\
w/o Ut.-Action    & \underline{89.7} & 49.1 & \underline{47.5} & 45.9 \\
w/o Int.-Decrease    & 83.9 & \underline{54.7} & 43.9 & 44.4 \\
w/o Ut.-Increase    & 86.2 & 39.6 & 47.3 & 43.5 \\
w/o Gate   & 86.2 & 47.2 & 46.7 & 43.0 \\ 
\hline
\textbf{AUSO} & \textbf{94.3} & \textbf{67.9} & \textbf{49.7} & \textbf{51.2} \\
\hline
\end{tabular}
\label{tab:alfworld_webshop}
\end{table}

\begin{table}[t]
\centering
\renewcommand{\arraystretch}{1.2}
\caption{Ablation Study for Period time for Internalize/Explore/Utilize in ALFWorld.}
\begin{tabularx}{\linewidth}{l>{\centering\arraybackslash}X>{\centering\arraybackslash}X}
\hline
Variant & \multicolumn{2}{c}{ALFWorld} \\
& ID Avg. & OOD Avg. \\
\hline
Ratio (3:5:2) & 85.1 & 62.3 \\
Ratio (3.3:3.3:3.3) & 87.4 & 41.5 \\
\hline
\textbf{AUSO (2:5:3)} & \textbf{94.3} & \textbf{67.9} \\
\hline
\end{tabularx}
\label{tab:alfworld}
\end{table}

\subsubsection{Baselines}

We compare AUSO with a diverse spectrum of agent learning methods. 
(1) \textbf{Prompt-based methods}: zero-shot and few-shot prompting, which directly condition the backbone model on task instructions without additional agent training \cite{brown2020language}. 
(2) \textbf{Prompt-based agentic and memory-based methods}: ReAct \cite{yao2023react}, Reflexion \cite{shinn2023reflexion}, ExpeL \cite{zhao2024expel}, Mem0 \cite{chhikara2025mem0}, and SimpleMem \cite{liu2026simplemem}, which improve multi-step decision making through in-context reasoning, verbal feedback, experiential memory, or external memory retrieval without updating the policy parameters. 
(3) \textbf{RL-based methods}: RLOO \cite{ahmadian2024back} and GRPO \cite{shao2024deepseekmath}, which optimize LLM policies with outcome rewards and group-relative or leave-one-out advantage estimation. 
(4) \textbf{Memory-augmented RL methods}: MemRL \cite{zhang2026memrl} and Memory-R1 \cite{yan2026memoryr1}, which integrate reinforcement learning with persistent or learnable memory mechanisms. 
(5) \textbf{Skill-augmented RL methods}: SkillRL \cite{xia2026skillrl}, Skill1 \cite{shi2026skill1}, SKILL0 \cite{lu2026skill0}, LatentSkill \cite{yu2026latentskill}, and Skill0.5 \cite{zhu2026skill05}, which represent recent attempts to train agents with external skills, internalized skills, or hybrid skill utilization and internalization.

\subsubsection{Implementation Details}
All experiments are conducted on a server equipped with \textbf{4 NVIDIA H200 GPUs}. Following Skill0.5~\cite{zhu2026skill05}, we adopt \textbf{Qwen2.5-7B-Instruct}~\cite{yang2024qwen25} as the backbone policy model for all experiments. For a fair comparison, all methods use the same backbone model, rollout configuration, maximum interaction length, and optimization budget within each benchmark. Specifically, for each task $q$, the policy performs $G=8$ rollouts to construct a rollout group for group-based policy optimization. Across all benchmarks, including ALFWorld, WebShop, and SearchQA, we train each method for \textbf{150 optimization steps}. For ALFWorld and WebShop, we follow the category-based ID/OOD evaluation protocol and report both in-distribution (ID) validation performance and out-of-distribution (OOD) generalization performance. For SearchQA, which does not adopt an ID/OOD split, we report task success following Skill0 to evaluate. More training and evaluation details are provided in the supplemental materials.


\subsection{Main Results}
Tables~\ref{tab:webshop_alfworld} and~\ref{tab:search_qa} report the main results on WebShop, ALFWorld, and Search-QA. Overall, \textbf{AUSO consistently achieves the strongest performance across heterogeneous agent scenarios}. More importantly, its improvements are observed under both ID and OOD settings, demonstrating effective skill utilization and generalization beyond seen tasks and unseen tasks.

\begin{figure}
    \centering
    \includegraphics[width=\linewidth]{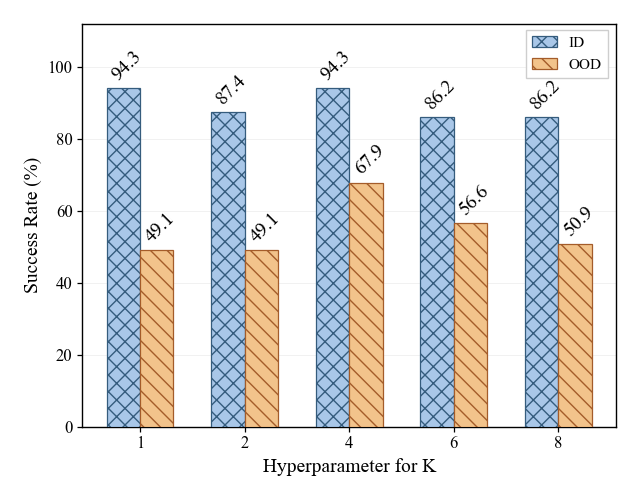}
    \caption{The ID and OOD success rate for different hyperparameter $K$ in the ALFWorld benchmark.}
    \label{fig:alfworld}
\end{figure}

\paragraph{WebShop}
On WebShop, AUSO achieves the best average performance under both ID and OOD settings, reaching \textbf{49.7} and \textbf{51.2}, respectively. Compared with Skill0.5, AUSO improves the two averages by \textbf{9.3} and \textbf{10.6} points. The gains are particularly pronounced on Electronics and Other under ID, as well as Access and Beauty under OOD. Notably, the larger improvement under OOD suggests that AUSO does not simply fit task-specific skill patterns. Instead, action-level optimization enables the agent to internalize useful skill knowledge and adapt its utilization according to the benefit of skill guidance for each action, resulting in stronger generalization to the unseen (OOD) tasks.

\paragraph{ALFWorld}
On ALFWorld, AUSO achieves an ID average of \textbf{94.3}, outperforming Skill0.5 by \textbf{1.2} points and obtaining the best overall performance. More importantly, AUSO reaches an OOD average of \textbf{67.9}, substantially surpassing Skill0.5 (\textbf{58.5}) by \textbf{9.4} points. The improvement is especially clear on challenging unseen tasks such as \textit{Pick2}, where the score increases from 33.3 to \textbf{54.2}. These results demonstrate that action-level skill utilization optimization facilitates the acquisition of transferable behavioral knowledge rather than merely fitting seen task configurations.

\paragraph{Search-QA}
The effectiveness of AUSO further extends to the knowledge-intensive reasoning tasks. AUSO achieves the highest overall average of \textbf{47.5}, outperforming SkillRL (\textbf{47.1}), Skill1 (\textbf{44.7}), and Skill0.5 (\textbf{44.2}). It achieves the best performance on TriviaQA (\textbf{64.7}) and HotpotQA (\textbf{44.0}), while remaining competitive across other single-hop and multi-hop tasks. These results demonstrate that AUSO generalizes beyond interactive environments and transfers action-level skill knowledge to diverse reasoning tasks.

\paragraph{Training and Generalization Analysis}
Fig.~\ref{fig:search} further compares the learning dynamics of AUSO and Skill0.5 on Search-QA. While the two methods exhibit comparable training success rates, AUSO achieves stronger validation performance as training progresses, particularly on multi-hop tasks where Skill0.5 shows noticeable fluctuations. This suggests that the advantage of AUSO does not simply come from better fitting training trajectories; instead, action-level optimization promotes more effective skill internalization and utilization, leading to improved generalization.

\subsection{Ablation Study}
We conduct comprehensive ablation studies on ALFWorld and WebShop to investigate the contribution of each component in AUSO. As shown in Table~\ref{tab:alfworld_webshop}, removing any component consistently degrades the overall performance, demonstrating that different stages of action-level skill optimization are complementary. In particular, removing \textit{Int.-Action} (Action-Level Internalization) leads to a substantial drop on ALFWorld, from \textbf{94.3/67.9} to 85.1/54.7 in ID/OOD settings, highlighting the importance of skill internalization. Similarly, removing \textit{Ut.-Action} (Action-Level Utilization) or the adaptive increase/decrease strategies in utilization/internalization results in consistent degradation, indicating that dynamically controlling skill utilization is crucial for transferring learned skills to different interaction states. Removing the gating mechanism also causes clear performance drops on both benchmarks, further validating its role in coordinating action-level skill optimization.

We further investigate the period allocation among different optimization stages in Table~\ref{tab:alfworld}. The adopted ratio of \textbf{2:5:3} achieves the best ID and OOD performance, suggesting that an appropriate balance between skill internalization, exploration, and utilization is important for effective policy optimization. Finally, Fig.~\ref{fig:alfworld} studies the influence of hyperparameter $K$. Performance peaks at $K=4$, achieving \textbf{94.3} ID and \textbf{67.9} OOD success rates, while either smaller or larger values lead to degradation, especially under OOD evaluation. This indicates that a moderate $K$ provides an effective balance for skill optimization and generalization.

\section{Conclusion}
In this paper, we proposed AUSO, an information gain based action-level via JSD unified skill optimization framework that unifies skill internalization and skill utilization for LLM agents. AUSO progressively learns from external skill guidance, consolidates autonomous decision-making through reinforcement learning, and finally uses action-level skill-sensitive signals to emphasize decisions that benefit from skills while suppressing those where skill guidance is unhelpful. Experiments on ALFWorld, WebShop, and SearchQA demonstrate that AUSO consistently outperforms competitive baselines and achieves stronger out-of-distribution generalization. Further ablation studies verify the contribution of progressive scheduling, action-level optimization, and the uncertainty-based gate. These results show that modeling skills at the action level provides an effective way to transform external guidance into transferable policy knowledge for long-horizon llm agents.


\bibliographystyle{ACM-Reference-Format}
\bibliography{sample-base}

\appendix

\section{Training Algorithm}

\subsection{Analysis of the Global Uncertainty Gate}
\label{app:competence_gate}

\paragraph{Statistical Justification of the Uncertainty Gate.} We provide a statistical interpretation of the global uncertainty gate
$g(p_q)=K \cdot p_q(1-p_q)$ used in the skill utilization stage of AUSO.
The key intuition is that action-level credit assignment is more reliable
when both successful and failed trajectories are sufficiently represented
within the rollout group. For a task $q$, suppose the rollout group contains $G$ trajectories, with
empirical success rate $p_q$. The expected numbers of successful trajectories $G_{+}$ and failed
trajectories $G_{-}$ are therefore be represented as:
\begin{equation}
    G_{+}=Gp_q,
    \qquad
    G_{-}=G(1-p_q)
\end{equation}

Let $I$ denote the action-level skill information gain score. To determine whether skill-sensitive actions are associated with successful outcomes, consider the
difference between the expected information scores of successful and failed
trajectories:
\begin{equation}
    \Delta_I
    =
    \mathbb{E}[I\mid R=1]
    -
    \mathbb{E}[I\mid R=0].
\end{equation}
Its empirical estimator is calculated as:
\begin{equation}
    \widehat{\Delta}_I
    =
    \overline{I}_{+}
    -
    \overline{I}_{-},
\end{equation}
where $\overline{I}_{+}$ and $\overline{I}_{-}$ denote the average
action-level information scores computed from successful and failed
trajectories, respectively.

Motivated by the standard variance analysis for the difference between two
sample means~\cite{welch1947generalization}, we analyze the statistical
reliability of this success--failure information contrast. To this end, we
consider the conditional variances of the action-level information score
under successful and failed outcomes:
\begin{equation}
    \sigma_{I,+}^{2}
    =
    \operatorname{Var}(I\mid R=1),
    \qquad
    \sigma_{I,-}^{2}
    =
    \operatorname{Var}(I\mid R=0)
\end{equation}
The variance of the estimated information contrast can then be characterized
as:
\begin{equation}
\begin{aligned}
    \operatorname{Var}(\widehat{\Delta}_{I})
    &=
    \frac{\sigma_{I,+}^{2}}{n_{+}}
    +
    \frac{\sigma_{I,-}^{2}}{n_{-}} \\
    &=
    \frac{\sigma_{I,+}^{2}}{Gp_q}
    +
    \frac{\sigma_{I,-}^{2}}{G(1-p_q)}
\end{aligned}
\label{eq:contrast_variance_general}
\end{equation}

To isolate the effect of the rollout outcome composition from
task-dependent variations in the information-score variance, we consider
the homoscedastic approximation:
\begin{equation}
    \sigma_{I,+}^{2}
    \approx
    \sigma_{I,-}^{2}
    \approx
    \sigma_I^{2}
\label{eq:homoscedastic_approximation}
\end{equation}
Under this approximation, Eq.~\eqref{eq:contrast_variance_general} becomes:
\begin{align}
    \operatorname{Var}(\widehat{\Delta}_{I})
    &\approx 
    \frac{\sigma_I^{2}}
    {G p_q(1-p_q)}
\label{eq:contrast_variance}
\end{align}

For a fixed rollout budget $G$ and information-score variance
$\sigma_I^{2}$, the precision of the estimated success--failure contrast,
defined as the inverse of its variance, therefore satisfies:
\begin{equation}
\begin{aligned}
    \operatorname{Precision}(\widehat{\Delta}_{I})
    &=
    \frac{1}
    {\operatorname{Var}(\widehat{\Delta}_{I})}\propto p_q(1-p_q)
\end{aligned}
\label{eq:contrast_precision}
\end{equation}
This result shows that the reliability of the success--failure contrast
depends directly on the composition of rollout outcomes. When $p_q$
approaches either $0$ or $1$, one outcome group becomes absent or severely
underrepresented, making the contrast increasingly unreliable.


\paragraph{Boundedness and Direction Preservation.}
Recall that the action-level modulation weight is:
\begin{equation}
    w_t
    =
    1+\beta(s)g(p_q)m_t
\end{equation}
where $|m_t|\leq1$ and $0\leq\beta(s)<1$. Since
$0\leq g(p_q)\leq1$, we have:
\begin{equation}
    1-\beta(s)
    \leq
    w_t
    \leq
    1+\beta(s)
\label{eq:app_weight_bound}
\end{equation}
Therefore, $w_t>0$, and the resulting action-level advantage
\begin{equation}
    A_{q,t}^{\mathrm{AUSO}}
    =
    w_t A_q^{\mathrm{GRPO}}
\end{equation}
preserves the sign of the original GRPO advantage:
\begin{equation}
    \operatorname{sign}
    \left(
        A_{q,t}^{\mathrm{AUSO}}
    \right)
    =
    \operatorname{sign}
    \left(
        A_q^{\mathrm{GRPO}}
    \right)
\label{eq:app_sign_preservation}
\end{equation}
Moreover,
\begin{equation}
    A_q^{\mathrm{GRPO}}=0
    \quad\Longrightarrow\quad
    A_{q,t}^{\mathrm{AUSO}}=0
\end{equation}
so the action-level information signal cannot introduce an artificial
optimization direction unsupported by the original outcome reward.

Finally, because the action-level modulation is centered within each task,
$\mathbb{E}_{t\in q}[m_t]=0$, we obtain:
\begin{equation}
\begin{aligned}
    \mathbb{E}_{t\in q}[w_t]
    &=
    1+
    \beta(s)g(p_q)
    \mathbb{E}_{t\in q}[m_t] =1
\end{aligned}
\label{eq:app_mean_preservation}
\end{equation}
Therefore, the global uncertainty gate does not alter the average
optimization scale within a task. Instead, it controls the reliability of
action-level modulation according to the available success--failure
contrast, while $m_t$ redistributes the resulting credit across the individual
actions of agent in the same trajectory.

\subsection{A Unified View of Internalization and Utilization}
\label{app:unified_objective}

\paragraph{Analysis of the Unified Procedure.}

Although skill internalization and skill utilization play different roles
during training, AUSO implements both through the same action-level
counterfactual information operator. The distinction between the two stages
lies not in how action-level information is measured, but in how the resulting
information signal is used for optimization. For an action $a_{q,t}$ taken at interaction history $h_{q,t}$, let
$C_{q,t}^{+}$ and $C_{q,t}^{-}$ denote two counterfactual context conditions.
We define the unified action-level information operator for the task $q$ as:
\begin{equation}
\mathcal{I}_{q,t}
\left(
C_{q,t}^{+},C_{q,t}^{-}
\right)
=
\operatorname{JSD}
\left(
\pi_{\theta}
\left(
\cdot\mid h_{q,t},C_{q,t}^{+}
\right)
\,\middle\|\,
\pi_{\theta}
\left(
\cdot\mid h_{q,t},C_{q,t}^{-}
\right)
\right)
\label{eq:unified_information_operator}
\end{equation}

Since one environment action may contain multiple generated tokens, the
implemented action-level score averages token-level divergences over the
action span:
\begin{equation}
\begin{aligned}
\mathcal{I}_{q,t}
&=
\frac{1}{|a_{q,t}|}
\sum_{j=1}^{|a_{q,t}|}
\operatorname{JSD}\Bigg(
\pi_{\theta}\!\left(
\cdot\mid h_{q,t},a_{q,t,<j},C_{q,t}^{+}
\right)
\\
&\qquad\qquad\qquad\quad
\Big\|
\pi_{\theta}\!\left(
\cdot\mid h_{q,t},a_{q,t,<j},C_{q,t}^{-}
\right)
\Bigg)
\end{aligned}
\label{eq:unified_action_jsd}
\end{equation}

The two learning stages instantiate this operator with different
counterfactual contexts:
\begin{equation}
\left(
C_{q,t}^{+},C_{q,t}^{-}
\right)
=
\begin{cases}
\left(
C_q^{\mathrm{general}}\cup C_q^{\mathrm{specific}},
C_q^{\mathrm{specific}}
\right),
& x\in\mathcal{H},
\\[4pt]
\left(
C_q^{\mathrm{specific}},
\varnothing
\right),
& q\in\mathcal{E}.
\end{cases}
\label{eq:counterfactual_context_pairs}
\end{equation}
Here, $\mathcal{H}$ and $\mathcal{E}$ denote the task regimes corresponding
to skill internalization and skill utilization, respectively.

For tasks in $\mathcal{H}$, the information score measures the discrepancy
between a privileged skill-conditioned teacher and the student. It is
therefore used as a direct supervision signal:
\begin{equation}
\mathcal{L}_{\mathrm{int}}(q)
=
\frac{1}{T_q}
\sum_{t=1}^{T_q}
w_{q,t}^{\mathrm{int}}
\mathcal{I}_{q,t}
\left(
C_q^{\mathrm{general}}\cup C_q^{\mathrm{specific}},
C_q^{\mathrm{specific}}
\right)
\label{eq:unified_internalization}
\end{equation}

For tasks in $\mathcal{E}$, the same operator instead measures how strongly
the current policy depends on task-specific skill conditioning:
\begin{equation}
\mathcal{I}_{q,t}^{\mathrm{util}}
=
\mathcal{I}_{q,t}
\left(
C_q^{\mathrm{specific}},
\varnothing
\right)
\label{eq:unified_utilization_information}
\end{equation}
This information is not optimized directly. Instead, it is converted into
the action-level modulation:
\begin{equation}
w_{q,t}^{\mathrm{util}}
=
1+
\beta(s)g(p_q)m_{x,t},
\label{eq:unified_utilization_weight}
\end{equation}
which reweights the original GRPO advantage:
\begin{equation}
\mathcal{L}_{\mathrm{util}}(q)
=
\mathcal{L}_{\mathrm{GRPO}}
\left(
q;
w_{q,t}^{\mathrm{util}}A_q
\right)
\label{eq:unified_utilization}
\end{equation}

Therefore, the unified procedure can be summarized as:
\begin{equation}
\mathcal{I}_{q,t}
\longrightarrow
\begin{cases}
\text{direct skill supervision},
& q\in\mathcal{H},
\\
\text{GRPO credit modulation},
& q\in\mathcal{E}.
\end{cases}
\label{eq:unified_information_usage}
\end{equation}
while tasks in the intermediate regime are optimized by standard GRPO without
additional skill-based correction.

\paragraph{Unified Routed Objective.}

Let $\rho_{\mathrm{int}}$, $\rho_{\mathrm{exp}}$, and
$\rho_{\mathrm{util}}$ denote the curriculum proportions allocated to
skill internalization, autonomous exploration, and skill utilization,
respectively:
\begin{equation}
\rho_{\mathrm{int}}
+
\rho_{\mathrm{exp}}
+
\rho_{\mathrm{util}}
=1.
\label{eq:curriculum_ratio}
\end{equation}

Rather than routing individual tasks to different objectives, AUSO
progressively changes the contribution of skill information along this
training curriculum. The complete objective is written as
\begin{equation}
\begin{aligned}
\mathcal{L}_{\mathrm{AUSO}}(\theta;s)
={}&
\mathcal{L}_{\mathrm{GRPO}}
\left(
\theta;
A_{q,t}^{\mathrm{GRPO}}
\left[
1+\beta(s)g(p_q)m_{q,t}
\right]
\right)
\\
&+
\lambda_0\alpha(s)
\sum_q
\mathbb{I}[p_q=0]
\mathcal{L}_{\mathrm{JSD}}^{(q)},
\end{aligned}
\label{eq:unified_curriculum_objective}
\end{equation}
where $\alpha(s)$ progressively decreases the contribution of
teacher-guided skill internalization, while $\beta(s)$ progressively
activates action-level skill utilization. The curriculum proportions
determine when these transitions occur, rather than routing individual
tasks into discrete optimization regimes.

Thus, GRPO remains the persistent optimization backbone throughout
training, while the role of skill information evolves progressively
from external skill internalization, through autonomous exploration,
to action-level information-gain-guided skill utilization.

\paragraph{Smooth reduction to the common backbone.}

The utilization branch continuously reduces to standard GRPO when the
action-level evidence is uninformative. Specifically,
\begin{equation}
w_{q,t}^{\mathrm{util}}=1
\end{equation}
whenever any of the following conditions holds:
\begin{equation}
p_q\in\{0,1\},
\qquad
m_{q,t}=0,
\qquad
\beta(s)=0.
\label{eq:utilization_reduction_conditions}
\end{equation}
Therefore, we can get the results under each of these conditions:
\begin{equation}
\mathcal{L}_{\mathrm{util}}(q)
=
\mathcal{L}_{\mathrm{GRPO}}(q)
\end{equation}

Furthermore, because:
\begin{equation}
\lim_{\beta_{(s)}\rightarrow 0}
w_{q,t}^{\mathrm{util}}
=
1,
\label{eq:weight_continuity}
\end{equation}
then we can get the following results:
\begin{equation}
\lim_{\beta_{(s)}\rightarrow 0}
\mathcal{L}_{\mathrm{util}}(q)
=
\mathcal{L}_{\mathrm{GRPO}}(q).
\label{eq:objective_continuity}
\end{equation}
Thus, utilization is not an independent or competing optimization objective;
it is a continuous, bounded refinement of the same GRPO objective. Similarly, when the internalization coefficient vanishes:
\begin{equation}
\lambda_{\mathrm{int}}(s)=0
\end{equation}
the hard auxiliary correction disappears. When both auxiliary mechanisms are inactive, we can get:
\begin{equation}
\mathcal{L}_{\mathrm{unified}}
\longrightarrow
\mathcal{L}_{\mathrm{GRPO}}
\label{eq:full_reduction_to_grpo}
\end{equation}

\paragraph{Unification result.}

Above establishes three forms of unification:

\begin{enumerate}
    \item \textbf{Signal unification:}
    internalization and utilization are both derived from same
    counterfactual action-level JSD operator;

    \item \textbf{Optimization unification:}
    all routing branches update a single policy under one mutually exclusive
    routed objective;

    \item \textbf{Backbone unification:}
    the proposed objective continuously reduces to standard GRPO when the
    auxiliary information is absent or unreliable.
\end{enumerate}

Hence, the framework does not combine unrelated auxiliary losses
heuristically. Instead, it instantiates a single counterfactual-information
principle in two complementary directions: reducing dependence on general
skills through internalization and preserving sensitivity to task-specific
skills through utilization.

\subsection{Why Jensen--Shannon Divergence for Action-Level Information Gain?}
\label{app:why_jsd}

\paragraph{Action-level information gain based on JSD}

At each environment state $h_t$, AUSO evaluates two counterfactual
executions of the same policy: one conditioned on the retrieved skill and
one without skill conditioning. Let $C\in\{0,1\}$ denote the skill context,
where $C=1$ represents skill-conditioned execution and $C=0$ represents
skill-free execution. Because the language-model action is generated autoregressively, we define the action distribution at the $\ell$-th response-token position as:
\begin{equation}
P_{t,\ell}(a)
=
\pi_\theta(a\mid h_{t,\ell},C=1),
\qquad
Q_{t,\ell}(a)
=
\pi_\theta(a\mid h_{t,\ell},C=0).
\end{equation}
Here, $h_{t,\ell}$ denotes the token-level history within environment
action $u_t$. In this paper, we assign a uniform prior over the two contexts:
\begin{equation}
p(C=0)=p(C=1)=\frac{1}{2}
\end{equation}
The corresponding marginal token distribution is
\begin{equation}
M_{t,\ell}(a)
=
p(a\mid h_{t,\ell})
=
\frac{1}{2}P_{t,\ell}(a)
+
\frac{1}{2}Q_{t,\ell}(a).
\end{equation}

The conditional mutual information between the skill context and the
token-level action is:
\begin{equation}
\begin{aligned}
I(C;A_{t,\ell}\mid h_{t,\ell})
&=
\sum_{c\in\{0,1\}}p(c)
D_{\mathrm{KL}}
\left(
p(A_{t,\ell}\mid h_{t,\ell},C=c)
\,\middle\|\,
p(A_{t,\ell}\mid h_{t,\ell})
\right)\\
&=
\frac{1}{2}
D_{\mathrm{KL}}
\left(
P_{t,\ell}\,\middle\|\,M_{t,\ell}
\right)
+
\frac{1}{2}
D_{\mathrm{KL}}
\left(
Q_{t,\ell}\,\middle\|\,M_{t,\ell}
\right)\\
&=
D_{\mathrm{JS}}
\left(
P_{t,\ell}\,\middle\|\,Q_{t,\ell}
\right)
\end{aligned}
\end{equation}
Therefore, we can get:
\begin{equation}
\boxed{
D_{\mathrm{JS}}
\left(
P_{t,\ell}\,\middle\|\,Q_{t,\ell}
\right)
=
I(C;A_{t,\ell}\mid h_{t,\ell})
}
\end{equation}
This identity shows that token-level JSD measures how much information
about the skill context is expressed in the local action distribution. In
particular, when:
\begin{equation}
D_{\mathrm{JS}}
\left(
P_{t,\ell}\,\middle\|\,Q_{t,\ell}
\right)
=0
\iff
P_{t,\ell}=Q_{t,\ell}
\end{equation}
which means that skill conditioning does not change the token-level
decision at that position.



\paragraph{Symmetry.}

The two distributions $P_{t,\ell}$ and $Q_{t,\ell}$ correspond to two
counterfactual decisions of the same policy. They do not form a
teacher--student pair with a predefined reference direction. A
one-sided KL divergence is direction-dependent:
\begin{equation}
D_{\mathrm{KL}}(P_{t,\ell}\|Q_{t,\ell})
\neq
D_{\mathrm{KL}}(Q_{t,\ell}\|P_{t,\ell})
\end{equation}
in general. Choosing one of these two directions would therefore introduce
an arbitrary asymmetry. In contrast, JSD satisfies
\begin{equation}
D_{\mathrm{JS}}(P_{t,\ell}\|Q_{t,\ell})
=
D_{\mathrm{JS}}(Q_{t,\ell}\|P_{t,\ell}),
\end{equation}
more suitable for comparing two counterfactual policy
decisions.

\paragraph{Boundedness.}

Because JSD is equal to the mutual information between a binary context
variable and the token action, it is bounded by the entropy of the
context:
\begin{equation}
0
\leq
D_{\mathrm{JS}}
(P_{t,\ell}\|Q_{t,\ell})
=
I(C;A_{t,\ell}\mid h_{t,\ell})
\leq
H(C)
=
\log 2.
\end{equation}
The action-level average is consequently also bounded:
\begin{equation}
0
\leq
\mathrm{IG}(u_t)
\leq
\log 2.
\end{equation}
This bounded scale is useful for stable action-level modulation. By
contrast, the KL divergence satisfies
\begin{equation}
D_{\mathrm{KL}}(P_{t,\ell}\|Q_{t,\ell})
\in[0,\infty]
\end{equation}
and may become arbitrarily large when $Q_{t,\ell}$ assigns very small
probability to an action favored by $P_{t,\ell}$.

\paragraph{Top-$k$ approximation used in implementation.}

Computing the exact JSD over the entire vocabulary is expensive. Therefore,
for each token position, AUSO constructs a coarse-grained support
\begin{equation}
\mathcal{S}_{t,\ell}
=
\operatorname{TopK}(P_{t,\ell})
\cup
\operatorname{TopK}(Q_{t,\ell})
\cup
\{\mathrm{tail}\}.
\end{equation}
The first two terms contain the union of the top-$k$ tokens selected by the
skill-conditioned and skill-free policies. All remaining vocabulary
probability mass is grouped into one aggregate tail category. Let
$\widetilde P_{t,\ell}$ and $\widetilde Q_{t,\ell}$ denote the resulting
normalized distributions on $\mathcal{S}_{t,\ell}$. The implemented
token-level score is then
\begin{equation}
\widehat{\mathrm{IG}}_{t,\ell}
=
D_{\mathrm{JS}}
\left(
\widetilde P_{t,\ell}
\,\middle\|\,
\widetilde Q_{t,\ell}
\right),
\end{equation}
and the implemented action-level score is
\begin{equation}
\widehat{\mathrm{IG}}(u_t)
=
\frac{1}{L_t}
\sum_{\ell}
m_{t,\ell}
\widehat{\mathrm{IG}}_{t,\ell}.
\end{equation}
The top-$k$ union and the aggregate tail preserve the normalization of both
distributions while avoiding the construction of two full
vocabulary-sized probability tensors.

\paragraph{Role in AUSO}

JSD is used as the unified action-level information signal throughout both internalization and utilization, providing a consistent measure of how skill knowledge influences individual decisions. During internalization, it quantifies the discrepancy between skill-conditioned teacher behavior and the skill-free counterfactual behavior, thereby identifying actions for which external skill guidance provides stronger corrective information. During utilization, the same measure characterizes how strongly the current policy decision depends on the retrieved skill, allowing the policy to selectively emphasize skill-sensitive actions. Therefore, the same action-level signal naturally connects the transition from external skill supervision to autonomous skill utilization, while providing symmetry, boundedness, support robustness, and a direct information-theoretic interpretation.

Importantly, $\mathrm{IG}(u_t)$ measures the influence of skill conditioning on the action distribution rather than directly representing a task-success probability. A large information gain indicates that introducing the skill substantially changes the policy's decision, but does not necessarily imply that this change is beneficial. The actual usefulness of such skill-induced changes is ultimately determined by the reinforcement-learning objective and the resulting task reward. In this way, JSD identifies \emph{where} skills exert meaningful influence, while environment feedback determines \emph{whether} that influence contributes to successful behavior.

\begin{figure}
    \centering
    \includegraphics[width=\linewidth]{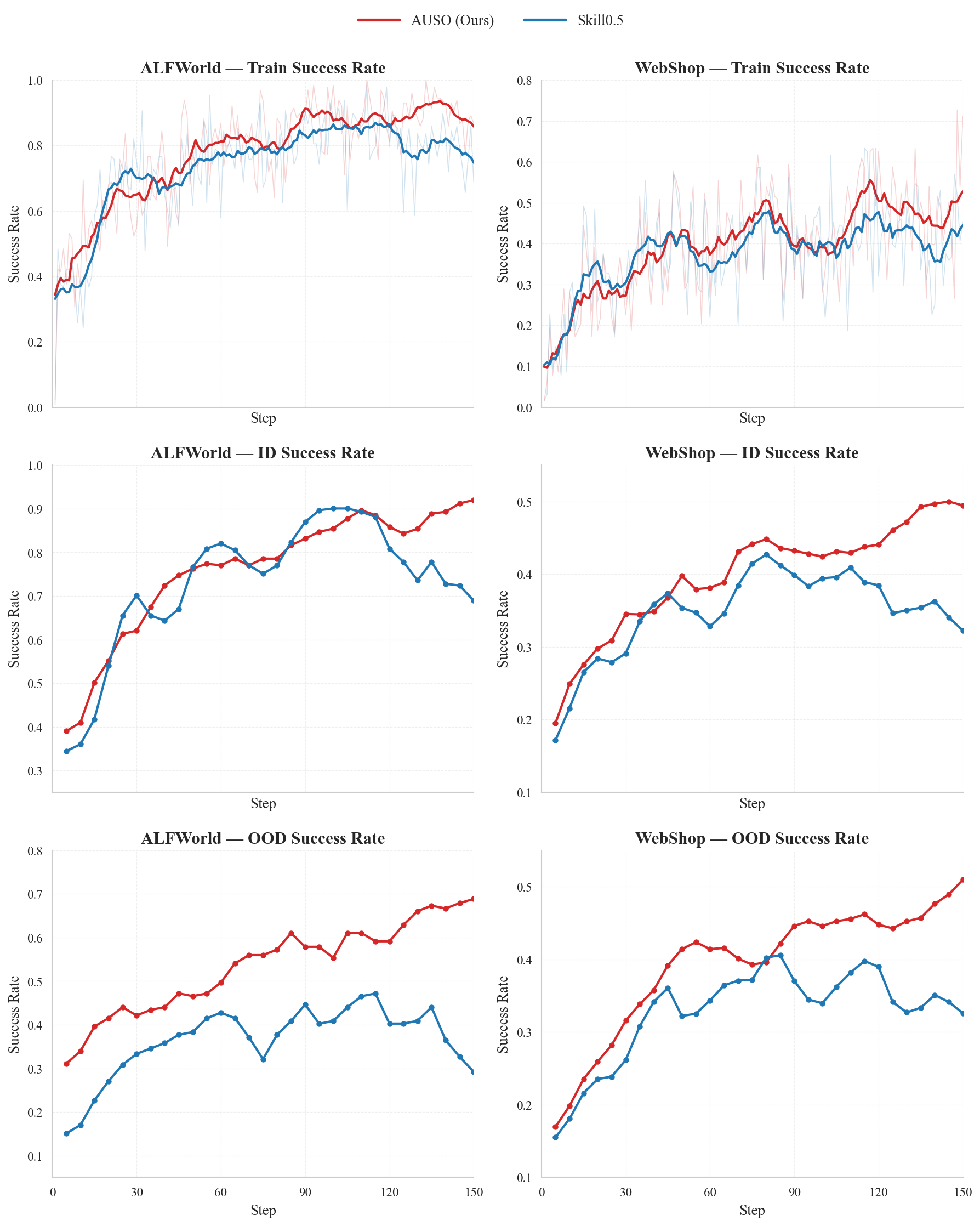}
    \caption{ID and OOD validation success rates and training success rates of our method and Skill0.5 with task-specific skills on Alfworld and Webshop benchmarks.}
    \label{supplementaw}
\end{figure}

\section{Experimental Setup}
\label{sec:experimental_setup}

\paragraph{Environments and ID/OOD protocol.}
We evaluate our method on ALFWorld, WebShop, and SearchQA. Across all benchmarks, policy optimization uses only in-distribution (ID) training tasks, while out-of-distribution (OOD) tasks are reserved for validation and final evaluation. ID and OOD performance is reported separately.

\paragraph{ALFWorld.}
Following Skill0.5, we treat \{Pick \& Place, Clean \& Place, Cool \& Place\}
as ID task types and \{Examine in Light, Heat \& Place, Pick Two \& Place\}
as OOD task types. Training samples are drawn only from the three ID types in
the standard training split. For evaluation, the standard
\texttt{eval\_in\_distribution} games are partitioned by task type into
disjoint ID and OOD subsets, and all available games are evaluated. ID tasks
use the ID skill bank, whereas OOD tasks use the corresponding OOD-specific
skill bank. We report per-task success rates and the macro-average within each
domain. The interaction horizon is 30 actions.

\paragraph{WebShop.}
We partition the 12,087 human-annotated goals into seven product domains. The
ID domains are \{Apparel, Electronics, Footwear, Other\}, and the OOD domains
are \{Accessories, Beauty \& Health, Home Decor\}. Following Skill0.5, we
downsample the dominant Other category using farthest-point sampling. This
produces 3,320 ID training goals, 454 ID evaluation goals, and 207 OOD
evaluation goals. Training uses only ID goals from the original training range
(indices 1,500 and above). The original test and development ranges (indices
0--1,499) are merged and then separated into ID and OOD evaluation subsets.
OOD goals never contribute to policy updates. We report exact task success,
per-category success rates, and ID/OOD averages. The interaction horizon is
15 actions.

\begin{figure}
    \centering
    \includegraphics[width=\linewidth]{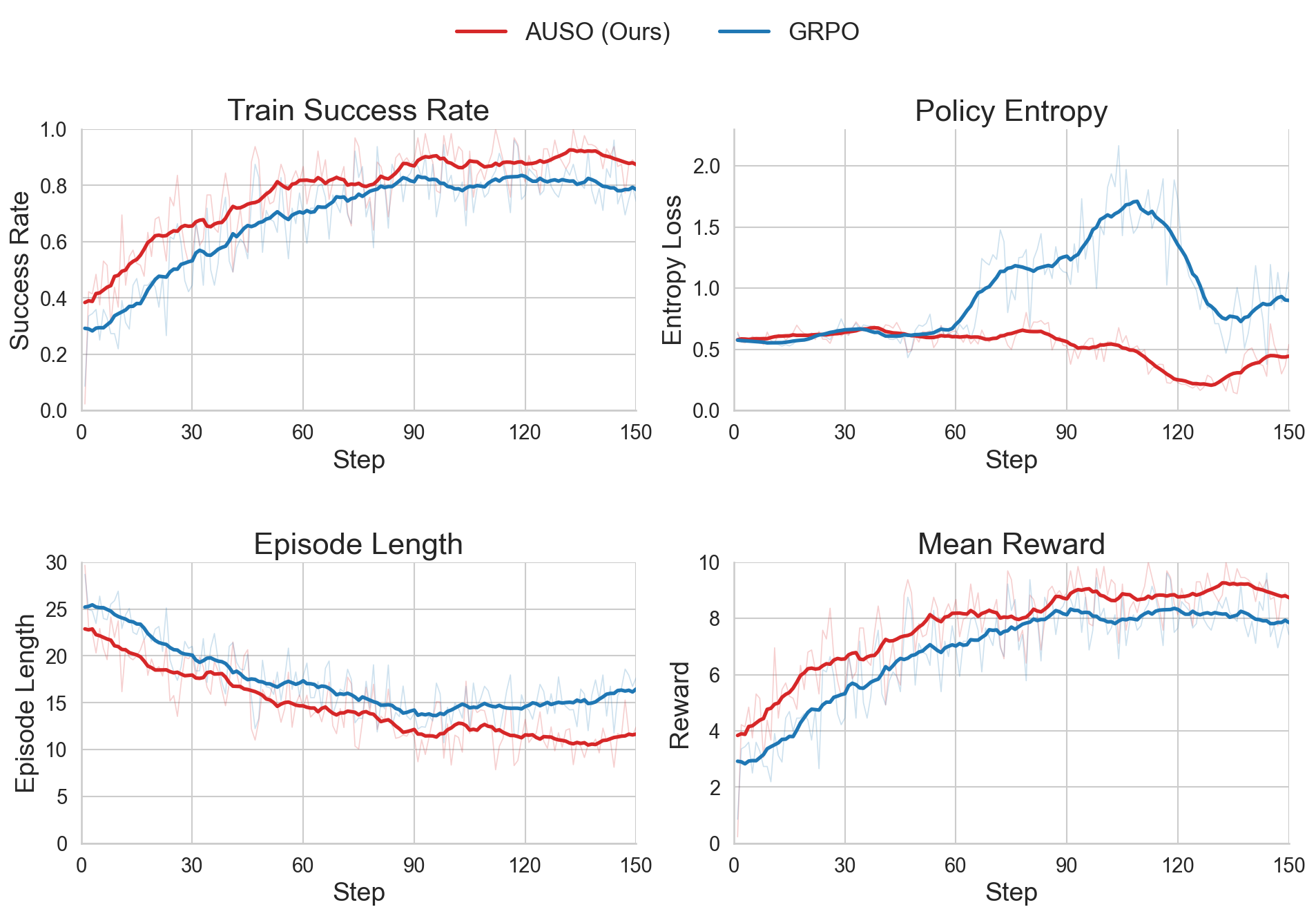}
    \caption{Training curves of AUSO vs. GRPO on ALFWorld: AUSO achieves higher success rate and reward with lower policy entropy and shorter episode length, indicating more confident and sample-efficient learning.}
    \label{supplementalfworld}
\end{figure}

\paragraph{SearchQA.}
The policy is trained only on the training splits of NQ and HotpotQA, which
form the ID group. Evaluation covers seven datasets: NQ and HotpotQA are ID,
while TriviaQA, PopQA, 2WikiMultiHopQA, MuSiQue, and Bamboogle are OOD. The
five OOD datasets are not used for policy optimization. We construct a
deterministic and domain-balanced validation set with random seed 42 and select the checkpoint at the last step. The selected checkpoint is then evaluated on the full test collection. All
datasets query the same Wikipedia corpus and complete SearchQA skill library;
ID/OOD labels are used only to identify training exposure and aggregate
metrics. We report per-dataset accuracy and ID/OOD averages. Each episode may
issue at most four search actions, with the top three passages returned per
query.

\paragraph{Implementation details.}
All experiments initialize from Qwen2.5-7B-Instruct and use GRPO with learning
rate \(1\times10^{-6}\), group size \(G=8\), and one policy epoch per update.
We train each benchmark for 150 optimization steps and validate every five
steps. The task batch sizes are 16, 16, and 128 for ALFWorld, WebShop, and
SearchQA, respectively. Skill banks remain fixed during training, allowing us
to evaluate policy learning and OOD skill utilization without online changes
to the external skill corpus.

\begin{table*}[t]
  \centering
  \small
  \renewcommand{\arraystretch}{1.5}
    \caption{Representative cases comparing AUSO and Skill0.5.}
  \setlength{\tabcolsep}{4pt}

  \resizebox{\textwidth}{!}{
  \begin{tabular}{llp{7.0cm}ccp{5.0cm}}
    \toprule
    \textbf{Domain} &
    \textbf{Method} &
    \textbf{Query / Task} &
    \textbf{\#Steps} &
    \textbf{Correct} &
    \textbf{Final output / action} \\
    \midrule

    \multirow{2}{*}{\shortstack{SearchQA\\(multi-hop)}}
      & AUSO (Ours)
      & \multirow{2}{7.0cm}{\textit{Frank Lamson-Scribner was adopted by a family near which town in Kennebec County?}}
      & 2
      & $\checkmark$
      & \texttt{Manchester} \\

      & Skill0.5
      &
      & 4
      & $\times$
      & \texttt{Manchester, Maine} \\

    \midrule

    \multirow{2}{*}{\shortstack{ALFWorld\\(OOD)}}
      & AUSO (Ours)
      & \textit{put a spraybottle in garbagecan}
      & 4
      & $\checkmark$
      & \texttt{put spraybottle 2 in/on garbagecan 1} \\

      & Skill0.5
      & \textit{put a cool lettuce in countertop}
      & 25
      & $\times$
      & \texttt{go to diningtable 1} \\

    \bottomrule
  \end{tabular}
  }
  \label{tab:bad_cases}
\end{table*}

\section{Detail Experimental Results}

\subsection{Training Dynamics of AUSO on ALFWorld}


As shown in Fig.~\ref{supplementaw}, we first compare the training dynamics of AUSO with Skill0.5 on ALFWorld and WebShop. To verify that the superior performance of AUSO on ALFWorld is not simply attributable to a longer training process, we extend Skill0.5 to the same training budget of 150 steps for a controlled comparison. Under this identical optimization budget, AUSO still achieves stronger performance, particularly on the ID and OOD evaluations, demonstrating that simply increasing the number of training steps is insufficient to match the gains brought by AUSO. As training proceeds, AUSO maintains or further improves its performance, whereas Skill0.5 tends to plateau or even degrade in the later stages despite continued optimization. This trend is especially evident under the more challenging OOD settings of both ALFWorld and WebShop, where AUSO exhibits a more stable improvement and the performance gap gradually becomes more pronounced. he consistent advantage on ID tasks indicates that the improvement in generalization does not come at the expense of skill-covered scenarios. These observations rule out additional training steps as the primary source of improvement and instead demonstrate that AUSO learns more effective and robust skill-conditioned behaviors under same training budget.

We further investigate the optimization behavior of AUSO against standard GRPO in Fig.~\ref{supplementalfworld}. AUSO achieves a consistently higher training success rate and mean reward throughout most of the optimization process, indicating that successful task behaviors are acquired earlier and reinforced more effectively. Meanwhile, AUSO exhibits lower policy entropy and progressively shorter episode lengths than GRPO, particularly during the middle and later stages of training. The reduced policy entropy suggests that the policy gradually forms more confident action distributions instead of remaining dependent on uncertain exploration, while the shorter trajectories indicate that AUSO learns to complete tasks with fewer redundant or ineffective interactions. Importantly, these improvements occur consistently across multiple optimization metrics rather than only in the final success rate. Taken together, the two analyses provide complementary evidence that the gains of AUSO originate from its more effective action-level skill optimization rather than a larger training budget, enabling the agent to acquire successful behaviors more efficiently, utilize skill information more effectively, and generalize more robustly to unseen task scenarios.

\subsection{Case Study: AUSO is More Sample-Efficient than Skill0.5}
\label{app:case_study}

\Needspace{18\baselineskip}


\begin{tcolorbox}[
    enhanced,
    colback=casegreenbg,
    colframe=casegreen,
    colbacktitle=casegreen,
    coltitle=white,
    fonttitle=\bfseries,
    title={AUSO --- Same Query [SUCCESS, 2 steps]},
    boxrule=0.9pt,
    arc=2mm,
    left=2mm,
    right=2mm,
    top=1mm,
    bottom=1mm
]
\textbf{Query:}
\textit{Frank Lamson-Scribner was adopted by a family near which town in
Kennebec County?}

\vspace{2pt}

\textbf{Step 0:}
\texttt{search: Frank Lamson-Scribner adopted family Kennebec County town}

\hfill
{\color{casegreen}\textbf{Success:} Answer-bearing passage retrieved}

\textbf{Observation:}
``was adopted by the Virgil Scribner family near Manchester, Maine''

\vspace{2pt}

\textbf{Step 1:}
\texttt{answer: Manchester}

\hfill
{\color{casegreen}\textbf{Success:} Minimal answer span; terminate}
\end{tcolorbox}

\vspace{3pt}

\begin{tcolorbox}[
    enhanced,
    colback=caseredbg,
    colframe=casered,
    colbacktitle=casered,
    coltitle=white,
    fonttitle=\bfseries,
    title={Skill0.5 --- Same Query [FAIL, 4 steps]},
    boxrule=0.9pt,
    arc=2mm,
    left=2mm,
    right=2mm,
    top=1mm,
    bottom=1mm
]
\textbf{Step 0:}
Retrieve the same answer-bearing passage.

\hfill
{\color{casegreen}\textbf{Success:} Sufficient evidence already available}

\textbf{Step 1:}
\texttt{search: manchester maine adopted by virgil scribner family}

\hfill
{\color{casered}\textbf{Failure:} Redundant search}

\textbf{Step 2:}
\texttt{search: manchester maine adopted by virgil scribner family near}

\hfill
{\color{casered}\textbf{Failure:} Irrelevant retrieval}

\textbf{Step 3:}
\texttt{answer: Manchester, Maine}

\hfill
{\color{casered}\textbf{Failure:} Over-answering under exact match}
\end{tcolorbox}

\paragraph{SearchQA (Multi-hop).}
For the same multi-hop SearchQA question, both AUSO and Skill0.5 successfully retrieve the answer-bearing passage in the first search step, indicating that their performance difference does not originate from initial retrieval quality. Given the sufficient evidence that Frank Lamson-Scribner ``was adopted by the Virgil Scribner family near Manchester, Maine,'' AUSO immediately identifies the relevant information, extracts the minimal answer span \texttt{Manchester}, and terminates the interaction after only two steps. In contrast, Skill0.5 fails to recognize that the retrieved evidence is already sufficient and continues issuing redundant follow-up queries about the adoption location. These unnecessary searches not only increase the interaction cost but also introduce additional and less relevant information into the reasoning context, eventually leading Skill0.5 to return the over-specified answer \texttt{Manchester, Maine}, which fails under exact-match evaluation. This comparison highlights that successful SearchQA reasoning depends not only on retrieving relevant evidence, but also on determining when the available evidence is sufficient and how it should be converted into a precise final answer. By making more effective action-level decisions, AUSO avoids unnecessary retrieval, terminates at an appropriate point, and produces a concise answer consistent with the required answer granularity. This case therefore illustrates how AUSO can translate retrieved evidence into more efficient and precise search-and-answer trajectories than Skill0.5.

\Needspace{20\baselineskip}

\begin{tcolorbox}[
    enhanced,
    colback=casegreenbg,
    colframe=casegreen,
    colbacktitle=casegreen,
    coltitle=white,
    fonttitle=\bfseries,
    title={AUSO --- ``Put a spraybottle in garbagecan'' [SUCCESS, 4 steps]},
    boxrule=0.9pt,
    arc=2mm,
    left=2mm,
    right=2mm,
    top=1mm,
    bottom=1mm
]
\textbf{Step 0:}
\texttt{go to countertop 1}

\hfill
{\color{casegreen}\textbf{Success:} Locate target object}

\textbf{Step 1:}
\texttt{take spraybottle 2}

\hfill
{\color{casegreen}\textbf{Success:} Acquire object}

\textbf{Step 2:}
\texttt{go to garbagecan 1}

\hfill
{\color{casegreen}\textbf{Success:} Navigate to target receptacle}

\textbf{Step 3:}
\texttt{put spraybottle 2 in/on garbagecan 1}

\hfill
{\color{casegreen}\textbf{Success:} Goal completed}
\end{tcolorbox}

\vspace{3pt}

\begin{tcolorbox}[
    enhanced,
    colback=caseredbg,
    colframe=casered,
    colbacktitle=casered,
    coltitle=white,
    fonttitle=\bfseries,
    title={Skill0.5 --- ``Put a cool lettuce in countertop'' [FAIL, 25 steps]},
    boxrule=0.9pt,
    arc=2mm,
    left=2mm,
    right=2mm,
    top=1mm,
    bottom=1mm
]
\textbf{Trajectory summary:}

\texttt{open fridge}
\(\rightarrow\)
\texttt{inspect countertop}
\(\rightarrow\)
\texttt{inspect cabinet}
\(\rightarrow\)
\texttt{revisit locations}
\(\rightarrow\)
\texttt{go to diningtable 1}

\vspace{2pt}

\textbf{Final observation:}
\texttt{Nothing happens}

\hfill
{\color{casered}\textbf{Failure:} Target lettuce still not found}

\vspace{2pt}

{\color{casered}\textbf{Failure:}}
Repeated exploration of already-checked locations.

\hfill
{\color{casered}\textbf{Failure:}}
Step budget exhausted.
\end{tcolorbox}

\paragraph{ALFWorld (OOD)}
A similar advantage is observed in the OOD ALFWorld case reported in Table~\ref{tab:bad_cases}. AUSO successfully completes the task \textit{``put a spraybottle in garbagecan''} within only four interaction steps. Specifically, it first navigates to the countertop to locate the target object, directly acquires the spraybottle, moves to the target garbagecan, and executes the required placement action. The entire trajectory remains compact and goal-directed, with each action making explicit progress toward task completion. In contrast, Skill0.5 fails to complete its OOD task even after exhausting the 25-step interaction budget. Its trajectory repeatedly explores previously inspected locations, including the fridge, countertop, and cabinet, without successfully locating the target lettuce or establishing a reliable direction for subsequent actions. Eventually, the agent moves to an unrelated location and terminates without satisfying the task objective. This substantial difference in trajectory efficiency suggests that AUSO is better able to translate its learned skills into effective action-level decisions in unseen scenarios, reducing redundant exploration and maintaining consistent progress toward the goal. Together with the SearchQA case, these examples demonstrate that AUSO improves not only final task success but also the efficiency and precision of intermediate decision making across different interactive environments.

Overall, these cases illustrate that AUSO not only improves task success but also produces more economical and goal-directed interaction trajectories. On SearchQA, it avoids unnecessary retrieval once sufficient evidence has been obtained and terminates with a concise answer, whereas redundant follow-up searches can introduce irrelevant information. On ALFWorld, AUSO follows a compact locate--acquire--navigate--place sequence instead of repeatedly revisiting ineffective states. These observations suggest that AUSO improves not only final task performance but also the efficiency and precision of intermediate decision making.

\begin{figure*}
    \centering
    \includegraphics[width=\linewidth]{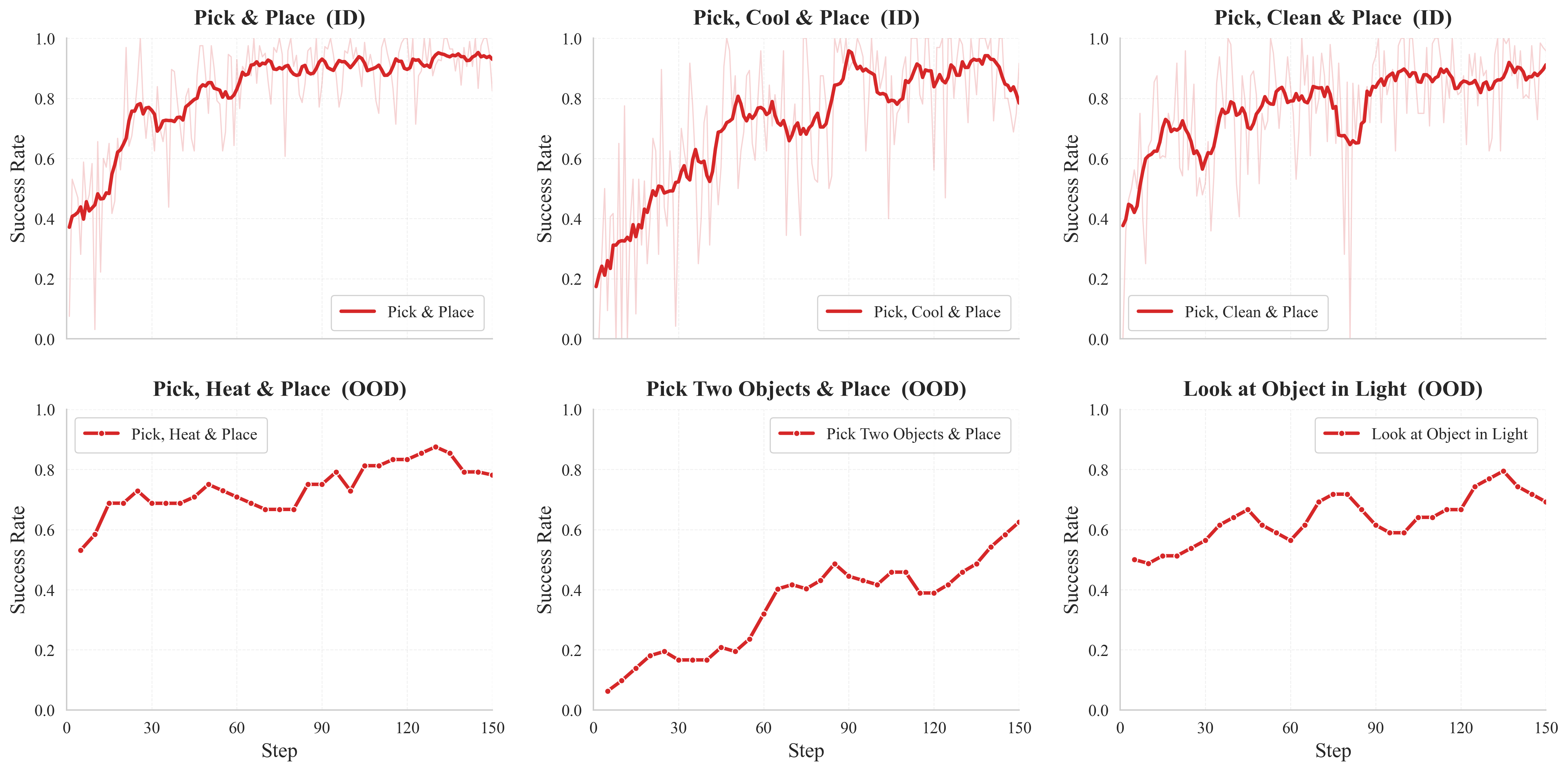}
    \caption{Success rate of each action on the ALFWorld benchmark for AUSO (ID for training part and OOD for validation)}
\end{figure*}

\begin{figure*}
    \centering
    \includegraphics[width=\linewidth]{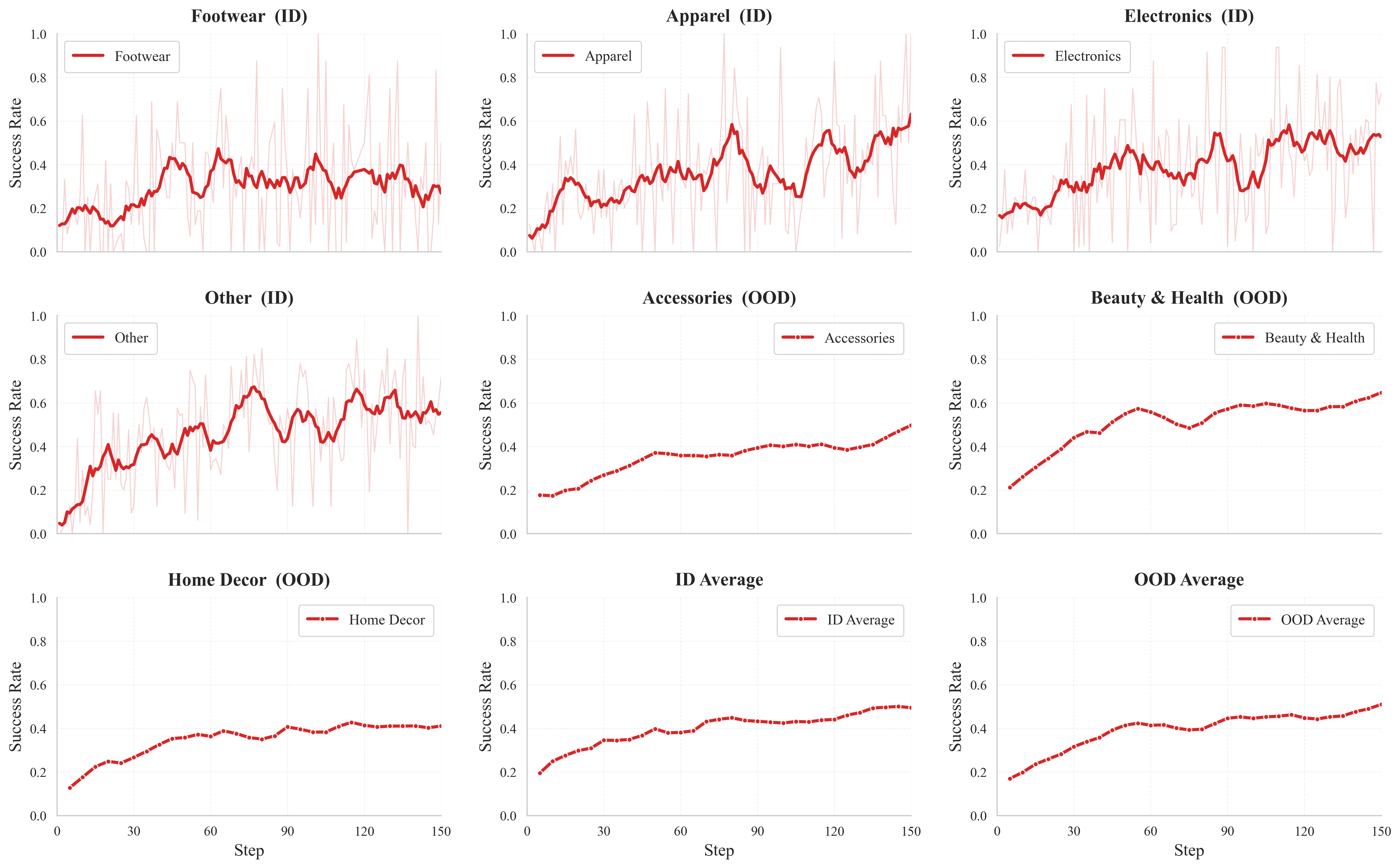}
    \caption{Success rate of each action on the Webshop benchmark for AUSO (ID for training part and OOD for validation)}
\end{figure*}

\begin{figure*}
    \centering
    \includegraphics[width=\linewidth]{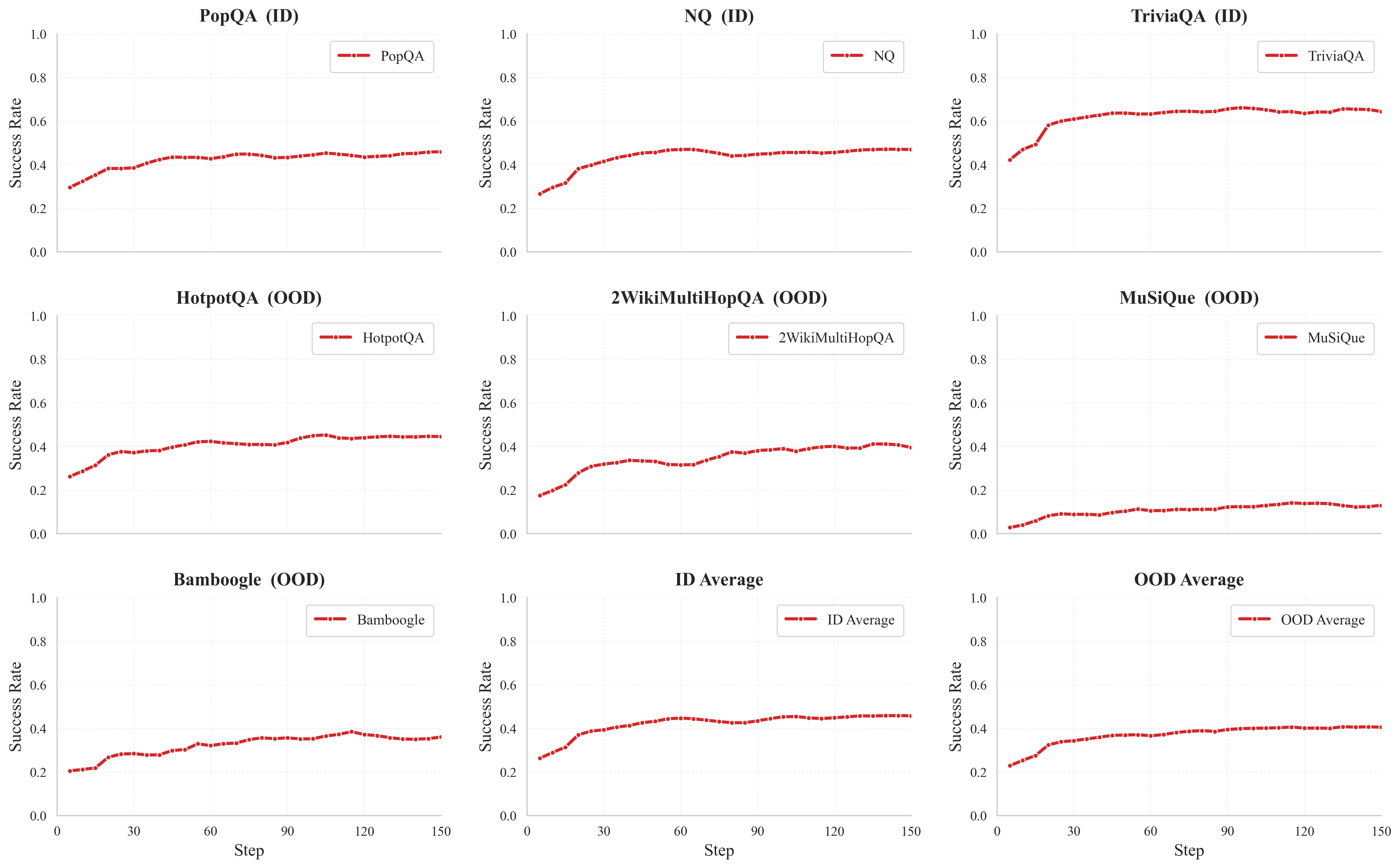}
    \caption{Success Rate of each Action in validation set in SearchQA benchmark for our AUSO method}
\end{figure*}

\end{document}